\documentclass[10pt,journal,compsoc]{IEEEtran}

\usepackage{graphicx}
\usepackage{amsmath}
\usepackage{amssymb}
\usepackage{algorithm}
\usepackage{graphics}
\usepackage{cite}
\usepackage{color}
\usepackage{ragged2e}
\usepackage{graphicx}
\usepackage{multirow}
\usepackage{booktabs}
\usepackage{makecell}
\usepackage{colortbl}
\usepackage[dvipsnames,table,xcdraw]{xcolor}
\usepackage{tikz}
\usepackage{pifont}
\usepackage[breaklinks=true,colorlinks, citecolor=black]{hyperref}
\usepackage{microtype}

\RequirePackage{silence}
\graphicspath{{./Imgs/},{./Imgs/authors/},{./imgs/Visualization/ade20k/},{./imgs/Visualization/cityscapes/}}

\DeclareGraphicsExtensions{.pdf,.jpg,.png}

\makeatletter
\DeclareRobustCommand\onedot{\futurelet\@let@token\@onedot}
\def\@onedot{\ifx\@let@token.\else.\null\fi\xspace}
\def\eg{\emph{e.g}\onedot} 
\def\ie{\emph{i.e}\onedot}

\def\etal{\emph{et al}\onedot}
\makeatother

\newcommand{\figref}[1]{Fig.~\ref{#1}}
\newcommand{\tabref}[1]{Tab.~\ref{#1}}
\newcommand{\secref}[1]{\S\ref{#1}}

\def\ie{\emph{i.e.}}
\def\eg{\emph{e.g.}}

\def\etal{{\em et al.~}}

\newcommand{\myPara}[1]{\vspace{6pt}\noindent\textbf{#1}}

\newcommand{\para}[1]{\vspace{0.05in}\noindent\textbf{#1}\quad}

\newcommand{\addFig}[1]{}
\newcommand{\addFigs}[1]{}

\def\ourmethod{LRFormer}

\begin{document}

\title{Low-Resolution Self-Attention for Semantic Segmentation}

\author{Yu-Huan Wu, Shi-Chen Zhang, Yun Liu, Le Zhang, Xin Zhan, Daquan Zhou,\\
Jiashi Feng, Ming-Ming Cheng, and Liangli Zhen
\IEEEcompsocitemizethanks{%

\IEEEcompsocthanksitem Y.-H.~Wu was with Nankai University when he started 
this research. 
Y.-H.~Wu and L. Zhen are currently with the Institute of High Performance Computing (IHPC), %
A*STAR, Singapore. (E-mail: wyh.nku@gmail.com, llzhen@outlook.com)
\IEEEcompsocthanksitem S.-C. Zhang, Y. Liu, and M.-M.~Cheng are with NKIARI, Shenzhen Futian and VCIP, Nankai University, Tianjin, China. 
(E-mail: zhangshichen@mail.nankai.edu.cn, liuyun@nankai.edu.cn, and cmm@nankai.edu.cn)
\IEEEcompsocthanksitem L. Zhang is with the University of Electronic Science 
and Technology of China. (E-mail: zhangleuestc@gmail.com)
\IEEEcompsocthanksitem X. Zhan is with Udeer AI, Hangzhou, China. 
(E-mail: zhanxin@udeer.ai)
\IEEEcompsocthanksitem D. Zhou is with School of Electronic and Computer Engineering, Peking University, China. (E-mail: zhoudaquan21@gmail.com)

\IEEEcompsocthanksitem J. Feng is with Bytedance Inc., Singapore. 
(E-mail: jshfeng@bytedance.com)
\IEEEcompsocthanksitem M.-M. Cheng is the corresponding author. %
}}

\IEEEtitleabstractindextext{%
\begin{abstract} 
\justifying{Semantic segmentation tasks naturally require
high-resolution information for pixel-wise segmentation
and global context information for class prediction. 
While existing vision transformers demonstrate promising performance,
they often utilize high-resolution context modeling,
resulting in a computational bottleneck.
In this work, we challenge conventional wisdom and 
introduce the Low-Resolution Self-Attention (LRSA) mechanism 
to capture global context at a significantly reduced computational cost, \ie, FLOPs.
Our approach involves computing self-attention in a fixed low-resolution space,
regardless of the input image's resolution, 
with additional $3\times 3$ depth-wise convolutions to capture fine details in the high-resolution space.
We demonstrate the effectiveness of our LRSA approach by building the LRFormer, 
a vision transformer with an encoder-decoder structure. 
Extensive experiments on the ADE20K, COCO-Stuff, and Cityscapes datasets 
demonstrate that LRFormer outperforms state-of-the-art models.
Code is available at \url{https://github.com/yuhuan-wu/LRFormer}.
}
\end{abstract}
}

\maketitle

\IEEEdisplaynontitleabstractindextext
\IEEEpeerreviewmaketitle

\section{Introduction}
\label{sec:intro}
As a fundamental computer vision problem,
semantic segmentation \cite{chen2017deeplab,fu2019dual,zhu2019asymmetric} 
aims to assign a semantic label to each image pixel.
Semantic segmentation models \cite{zhao2017pyramid,yang2018denseaspp} 
usually rely on pretrained backbone networks 
\cite{he2016deep,xie2017aggregated} for feature extraction, 
which is then followed by specific designs for pixel-wise predictions. 
In the last decade, the progress in feature extraction via various 
backbone networks
has consistently pushed forward state-of-the-art semantic segmentation
\cite{huang2023ccnet,xie2021segformer,yuan2021hrformer}. 
This paper improves the feature extraction for semantic segmentation 
from a distinct perspective.

It is commonly believed that semantic segmentation, 
as a dense prediction task, 
requires high-resolution features to ensure accuracy. 
In contrast, image classification typically infers predictions 
from a very small feature map, such as 1/32 of the input resolution.
Semantic segmentation models with convolutional neural networks (CNNs) 
usually decrease the strides of backbone networks to increase 
the feature resolution 
\cite{zhao2018psanet,zhang2018context,yu2018learning,takikawa2019gated},
\eg, 1/8 of the input resolution.
This attribute is also well preserved in transformer-based 
semantic segmentation, 
demonstrating that high-resolution is still necessary 
for semantic segmentation.

\begin{figure}[!t]
  \centering
  \includegraphics[width=.8\linewidth]{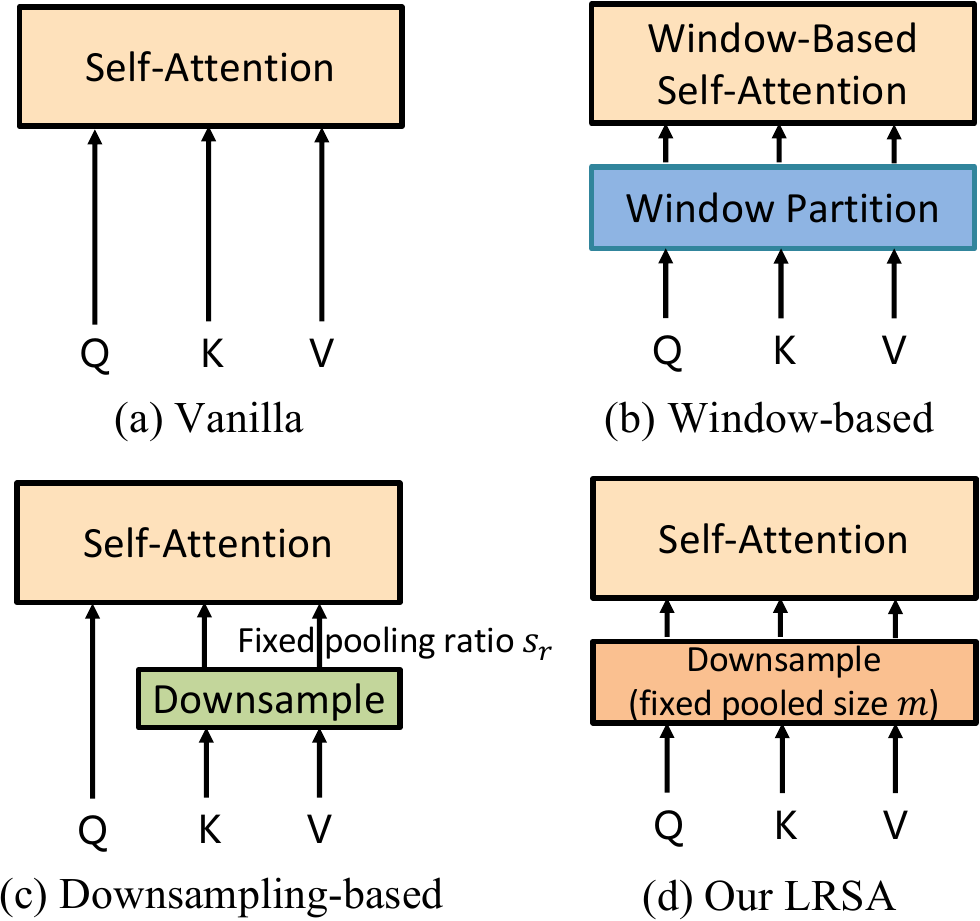}
  \caption{\textbf{Comparison of existing and our proposed paradigms 
    for the self-attention calculation in the vision transformer}.
    Representatives include (a) ViT \cite{dosovitskiy2021image}, 
    DeiT \cite{touvron2020training}; 
    (b) Swin \cite{liu2021swin}, CSwin \cite{dong2022cswin}; 
    (c) PVT \cite{wang2021pyramid}, SegFormer \cite{xie2021segformer}, 
    P2T \cite{wu2022p2t}; and (d) our \ourmethod. 
    Positional encoding modules are not drawn in this module.
    More details of (d) can refer to \figref{fig:basic_block}.
  }\label{fig:diff_our_att}
\end{figure}

High-resolution features are powerful for capturing the local details, 
while context information pertains to the broader understanding of the scene. 
Contextual features discern the interrelations between 
various scene components \cite{yuan2020object}, 
mitigating the ambiguity inherent in local features.
Thus, considerable research efforts \cite{yu2015multi,chen2017deeplab}
have been devoted to extending the receptive field of CNNs.
Conversely, vision transformers inherently facilitate the computation of global 
relationships by introducing self-attention with a global receptive field.
Nonetheless, this comes at a significant computational cost, 
as vanilla attention mechanisms exhibit quadratic complexity to input length.
Intriguingly, seminal studies \cite{xie2021segformer,wang2021pvtv2,wu2022p2t} 
made a remarkable effort by judiciously downsampling some of the features 
(\ie, key and value) during the self-attention computation for reduced 
computational complexities. 

Nevertheless, we observe that the computational overhead of self-attention 
remains a non-negligible bottleneck for existing vision transformers, 
as evidenced by \tabref{tab:memory_flops}. 
Consequently, we aim to delve deeper into the downsampling 
in the core component of the transformer, \ie, self-attention. 
Diverging from prior works that only downsample the key and value features 
\cite{xie2021segformer,wang2021pvtv2,wu2022p2t}, 
we propose to downsample all constituents—query, key, and value features. 
In this way, the output of self-attention would be in a low-resolution 
so that the mainstream of the transformer would contain low-resolution. 
Furthermore, we adopt a fixed downsampling size rather than 
a downsampling ratio to attain a very low computational complexity 
for  self-attention. 
The proposed method is called \textbf{Low-Resolution Self-Attention (LRSA)}.

\figref{fig:diff_our_att} depicts the differences between existing 
self-attention approaches and our LRSA.
Vanilla self-attention \cite{dosovitskiy2021image} 
(\figref{fig:diff_our_att}(a)) directly computes the global feature relations 
in the original resolution, which is quite expensive.
Window-based methods 
\cite{liu2021swin,liu2022swinv2,yang2021focal,dong2022cswin} 
(\figref{fig:diff_our_att}(b)) divide the features into small windows 
and perform local self-attention within each window.
Downsampling-based methods \cite{wang2021pyramid,xie2021segformer,fan2021multiscale,wu2022p2t} 
(\figref{fig:diff_our_att}(c)) keep the size of the query unchanged, 
and they downsample the key and value features with a fixed pooling ratio. The lengths of
key and value features increase linearly with the input resolution.
In contrast, our LRSA  (\figref{fig:diff_our_att}(d)) downsamples all query, 
key, and value to a small fixed size, 
leading to very low complexity regardless of the input resolution.
More analysis of the computational complexity can refer to \secref{sec:lrsa}.

\begin{figure}[!t]
  \centering
  \includegraphics[width=\columnwidth]{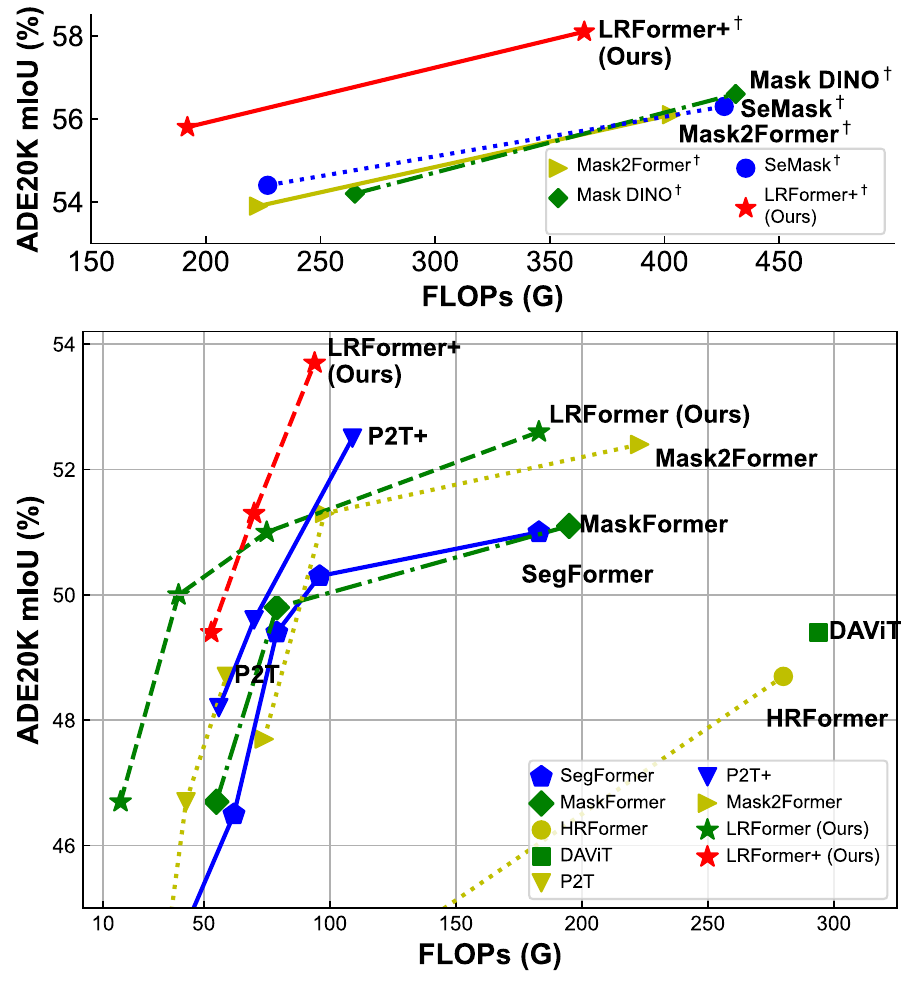}
  \caption{
    \textbf{Experimental comparisons on ADE20K 
    \cite{zhou2017scene} dataset. }
    Methods marked with ``$\dagger$'' are the results pretrained on the ImageNet-22K dataset. 
    Data are from \tabref{tab:exp_ade20k} and \tabref{tab:exp_ade20k_advanced}.
  }\label{fig:cmp_ade20k}
\end{figure}

While LRSA significantly boosts efficiency in capturing global context, 
we recognize that maintaining fine-grained details is another critical 
aspect for optimal performance in semantic segmentation.
To address this duality, we employ LRSA to capture global context information
in a purely low-resolution domain,
while simultaneously integrating small kernel (3$\times$3) depth-wise 
convolution to capture local details in the high-resolution space. 
Based on these foundational principles,
we build a new backbone network for feature extraction and a simple decoder 
to aggregate the extracted multi-level features for semantic segmentation. 
This new model is dubbed as \textbf{Low-Resolution Transformer (\ourmethod)}. 
We evaluate \ourmethod~on popular benchmarks, including 
ADE20K \cite{zhou2017scene}, 
COCO-Stuff \cite{caesar2018coco}, 
and Cityscapes \cite{cordts2016cityscapes}. 
Experimental results (\eg, \figref{fig:cmp_ade20k}) demonstrate 
the superiority of \ourmethod~series over state-of-the-art models.

\section{Related Work}

\subsection{Semantic Segmentation}

Semantic segmentation is a fundamental task in computer vision.
It is challenging due to the numerous variations like object sizes, textures, and lighting conditions in practical scenarios.
FCN \cite{long2015fully}, the pioneering work in this area, proposed the adaptation of CNNs for semantic segmentation in an end-to-end manner.
Since then, numerous studies have been built upon FCN \cite{long2015fully}, with major efforts focused on
 enriching multi-scale representations \cite{ zhao2017pyramid,chen2017rethinking,yang2018denseaspp}, enhancing boundary perception \cite{ding2019boundary, li2020improving, yuan2020segfix, zhen2020joint}, contextual representations \cite{yuan2018ocnet, yuan2020object} and introducing visual attention\cite{zhao2018psanet, zhu2019asymmetric, fu2019dual, takikawa2019gated, huang2023ccnet}.
These studies deeply explored the semantic head design upon FCN \cite{long2015fully} and achieved great progress.
Among these,
many approaches \cite{zhao2017pyramid,chen2017deeplab,fu2019dual,zhu2019asymmetric,huang2023ccnet,yang2018denseaspp,zhao2018psanet,zhang2018context,yu2018learning,takikawa2019gated, wu2021jcs,wang2024class,liang2024relative} 
are greatly benefited from the high-resolution features, performing prediction in the 1/8 of the input resolution to ensure high accuracy.
Moreover, there exist some query-based frameworks \cite{li2022panoptic, cheng2021per, cheng2022masked, cheng2022masked, zhang2021k, yu2022k,li2023mask} that changed the decoder style in semantic segmentation. 
For example, Panoptic SegFormer \cite{li2022panoptic} introduced a unified mask prediction workflow using a query decoupling strategy to handle thing and stuff categories separately.
MaskFormer \cite{cheng2021per} revolutionized mask decoders with transformer-based mask classification.
Mask2Former \cite{cheng2022masked} significantly enhanced MaskFormer via multi-scale masked attention.
K-Net \cite{zhang2021k} unifies various segmentation tasks through dynamic kernels.
K-means Transformer\cite{yu2022k}  integrated K-means clustering to optimize the mask generation for better instance

More recently, many works \cite{zheng2021rethinking, xie2021segformer, yuan2021hrformer, cheng2021per, cheng2022masked} showed that vision transformers \cite{dosovitskiy2021image} can largely improve the performance of semantic segmentation.
This is mainly attributed to the strong global capability of vision transformers, which happens to be a crucial property required for semantic segmentation.
For example, SETR \cite{zheng2021rethinking} first adapted ViT as an encoder followed by multi-level feature aggregation. 
SegFormer \cite{xie2021segformer} introduced a novel pyramid vision transformer encoder with an MLP mask decoder.
FeedFormer \cite{shim2023feedformer} introduced a transformer-based decoder with features as queries, significantly enhancing the structure information.
More discussions on vision transformers can refer to \secref{sec:vits}.

Recently, there exist some universal or foundational vision models exploring large scale capabilities, such as EVA \cite{fang2023eva}, OneFormer \cite{jain2023oneformer}, and One-Peace \cite{wang2023one}. 
They also achieved a great success in semantic segmentation due to their superior capability, some of which are even very capable to process multi-modal input.
For clarity, we are not able to compare our LRFormer with these large vision models due to limited GPU resources currently.

\subsection{Convolutional Neural Networks}
Given that CNN-based semantic segmentation models rely on CNN backbones for feature extraction, we discuss some notable CNN architectures.
Since the emergence of AlexNet \cite{krizhevsky2012imagenet}, 
many techniques have been developed to strengthen the CNN representations and achieved great success.
For example, VGG \cite{simonyan2014very}, GoogleNet \cite{szegedy2015going}, ResNets \cite{he2016deep} and DenseNets \cite{huang2017densely} developed increasingly deep CNNs to learn more powerful representations.
ResNeXts \cite{xie2017aggregated}, Res2Nets \cite{gao2019res2net}, and ResNeSts \cite{zhang2022resnest} explored the cardinal design in ResNets \cite{he2016deep}.
SENet \cite{hu2018squeeze} and SKNet \cite{li2019selective} introduced different attention architectures for selective feature learning.
Very recently, CNNs with large kernels are proven powerful in some works \cite{liu2022convnet,ding2022scaling,liu2022more}.
To ensure the high-resolution of feature maps for accurate semantic segmentation, semantic segmentation models usually decrease the strides of these CNNs and uses the dilated convolutions \cite{yu2015multi} to keep larger receptive field.
Motivated by this, HRNet \cite{wang2020deep} was proposed to directly learn high-resolution CNN features.
Despite the numerous successful stories, CNNs are limited in capturing global and long-range relationships, which are of vital importance for semantic segmentation.

\subsection{Vision Transformers}
\label{sec:vits}
Transformers are initially proposed in natural language processing (NLP) \cite{vaswani2017attention}.
Through multi-head self-attention (MHSA), transformers are capable of modeling global relationships.
Thanks to this characteristic,
transformers may also be powerful for computer vision tasks that require global information for a whole understanding of the visual scenarios.
To bridge this gap, ViT \cite{dosovitskiy2021image} transformed an image to tokens via a 16$\times$16 pooling operation and adopts the transformer to process these tokens, achieving better performance than CNNs in image recognition.
After that, vision transformers are developed rapidly by leveraging knowledge distillation \cite{touvron2021training}, overlapping patch embedding \cite{yuan2021tokens} or convolutions \cite{wu2021cvt,chu2021conditional}.
Recently, pyramid vision transformers \cite{wang2021pyramid,wang2021pvtv2,liu2021swin,yuan2021hrformer,wu2022p2t,fan2021multiscale, zhou2022understanding, yu2022metaformer} are proven to be powerful for image recognition tasks like semantic segmentation.
For example, 
PVT \cite{wang2021pyramid} and MViT \cite{fan2021multiscale} proposed to build a pyramid vision transformer pipeline via performing downsampling on key and value features.
Notably, MViT \cite{fan2021multiscale} decreases the resolution of the query by half in the first block of each stage, without the need of patch embedding between each stage.
Liu \etal \cite{liu2021swin} created a window-based vision transformer with shifted windows.
Yuan \etal \cite{yuan2021hrformer} presented HRFormer to learn high-resolution features for dense prediction using the vision transformer. 
Xia \etal \cite{xia2022vision} proposed DAT with deformable attention, conducting deformable sampling on key and value features.
Wu \etal \cite{wu2022p2t} introduced an efficient and multi-scale self-attention strategy via in-layer pyramid pooling.
Liu \etal \cite{liu2021vision} proposed to compute the self-attention in a hierarchical manner.
More approaches can refer to the survey \cite{guo2022attention}.

Despite their reported effectiveness, it is still commonly believed that high-resolution features are crucial for self-attention to effectively capture contextual information in semantic segmentation.
Window-based vision transformers \cite{liu2021swin,liu2022swinv2,yang2021focal,dong2022cswin} calculate self-attention within each local windows to reduce the computational complexity so that they can keep the high-resolution of feature maps. 
Downsampling-based vision transformers \cite{wang2021pyramid,wang2021pvtv2,xie2021segformer,fan2021multiscale, wu2022p2t, liu2021vision} keep the size of the query while partially conduct the downsampling on the key and value features with a fixed pooling ratio. Such strategy greatly reduces the complexity compared with vanilla attention so as to keep high-resolution features, making themselves computationally non-negligible especially for high-resolution inputs (\tabref{tab:memory_flops}). 
In contrast, we question the necessity of  keeping high-resolution for capturing context information via self-attention. 
We study this question by proposing LRFormer with LRSA. 
The good performance on several public benchmarks suggest the superiority of our LRFormer for semantic segmentation.

\section{Methodology}

In this section, we first introduce the Low-Resolution Self-Attention (LRSA) mechanism in \secref{sec:lrsa}.
Then, we build Low-Resolution Transformer (\ourmethod) using LRSA for semantic segmentation in \secref{sec:lrformer}. 
The decoder of \ourmethod~is presented in \secref{sec:decoder_head}.
Finally, we provide the implementation details in \secref{sec:implement_details}.

\begin{table}[!t]
  \centering
  \setlength{\tabcolsep}{1mm}
  \caption{\textbf{Comparison of various self-attention schemes.}
  $N$ is the length of the flattened features and $C$ is the number of feature channels. 
  ``Spatial Corr.'' denotes the spatial correlation.
  We omit constant factors for simplicity, like the window size in window-based methods and the downsampled size of LRSA.
  }
  \label{tab:cmp_attention_complexity}
\resizebox{\columnwidth}{!}{%
  \begin{tabular}{l|cccc} \Xhline{1pt}
      Scheme
      & Global & Spatial Corr. & Complexity \\
      \Xhline{1pt}
      Window-based \cite{liu2021swin} & \ding{56} & \ding{52} & $O(NC^2)$ \\
      Factorized \cite{xu2021coat} & \ding{52} & \ding{56} &  $O(NC^2)$ \\
      Downsampling-based \cite{xie2021segformer} & \ding{52} & \ding{52} & $O(N^2C+NC^2)$\\
      LRSA (Ours) & \ding{52} & \ding{52} & $O(NC+C^2)$ \\
      \Xhline{1pt}
  \end{tabular}}  
\end{table}

\subsection{Low-Resolution Self-Attention}
\label{sec:lrsa}

Unlike existing vision transformers that aim to maintain high-resolution feature maps during self-attention, 
our proposed LRSA computes self-attention in a low-resolution space, significantly reducing computational costs.
Before delving into our proposed LRSA, let us first revisit the vision transformer architecture.

\para{Revisiting self-attention in transformers.}
The vision transformer \cite{dosovitskiy2021image} has been demonstrated to be very powerful for computer vision \cite{wang2021pyramid,wang2021pvtv2,liu2021swin,yuan2021hrformer,wu2022p2t,fan2021multiscale,liu2022swinv2,yang2021focal,dong2022cswin}.
It consists of two main parts: the multi-head self-attention (MHSA) and the feed-forward network (FFN). We continue by elaborating on MHSA. Given the input feature $F_{in}$, the query $Q$, key $K$ and value $V$ are obtained with a linear transformation from $F_{in}$. Then, we can calculate the vanilla MHSA as
\begin{equation}
\label{eq:mhsa}
\text{Attention}(F_{in}) = \text{Softmax}(\frac{Q K^T}{\sqrt{d_k}})V,
\end{equation}
where $d_k$ is the number of channels of $F_{in}$. We omit the multi-head operation for simplicity. The overall computational cost of vanilla self-attention is $O(N^2C+NC^2)$, where $N$ and $C$ are the number of tokens and the number of channels of $F_{in}\in \mathbb{R}^{N\times C}$, respectively. As the number of tokens of natural images are usually very large, the computational cost of vanilla self-attention is very high.

\para{Previous solutions.}
To alleviate the computational cost while keeping the high-resolution of feature maps, recent downsampling-based vision transformers \cite{wang2021pyramid,wang2021pvtv2,xie2021segformer,fan2021multiscale,wu2022p2t} change the self-attention computation to
\begin{equation}\label{eq:pmhsa}
\text{Attention}(F_{in}) = \text{Softmax}(\frac{Q K_s^T}{\sqrt{d_k}})V_s,
\end{equation}
in which $K_s$ and $V_s$ are the downsampled key $K$ and value $V$ with a fixed downsampling ratio $s_r$, respectively.
The $1D\leftrightarrow 2D$ feature reshaping is omitted for convenience. The length of $K_s$ and $V_s$ is 1/$s_r^2$ of the original $K$ and $V$. If the original length of $K$ and $V$ is too large, the $K_s$ and $V_s$ will also be long sequences, introducing considerable computational cost in self-attention.
Here, we only introduce downsampling-based transformers because they are most relevant to our method.

\begin{figure}[!t]
  \centering
  \includegraphics[width=\linewidth]{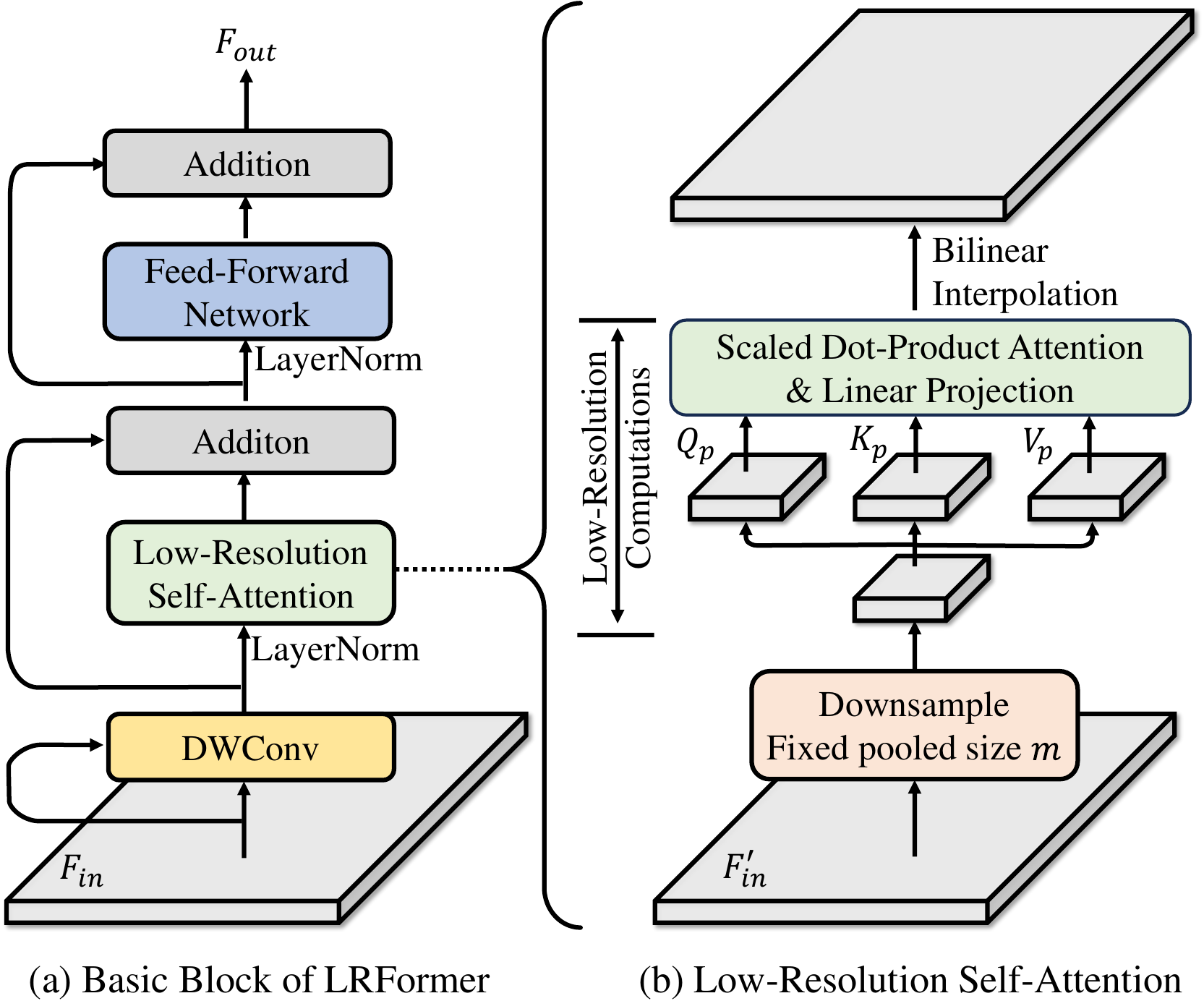}
  \caption{\textbf{Illustration of a basic block of our \ourmethod. }
  We add a 3$\times$3 depth-wise convolution (DWConv) with a residual connection before LRSA, which is also applied between the two linear layers of the FFN.
  }
  \label{fig:basic_block}
\end{figure}

\para{Our solution.}
Instead, we tackle the heavy computation of vanilla self-attention from a new perspective: we do not keep the high-resolution of feature maps but process the features in a very low-resolution space (\figref{fig:basic_block} (b)). 
Specifically, the proposed LRSA downsamples the input feature $F_{in}$ to a fixed size $m$. Then, multi-head self-attention is applied:
\vspace{-1mm}
\begin{equation}\label{eq:lrsa}
  \text{Attention}(F_{in}) = \text{Softmax}(\frac{Q_pK_p^T}{\sqrt{d_k}})V_p,
\vspace{-1mm}
\end{equation}
where $Q_p$, $K_p$ and $V_p$ are obtained by a linear transformation from the downsampled $F_{in}$. $Q_p$, $K_p$ and $V_p$ are with a fixed size $m$, regardless of the resolution of the input $F_{in}$.
Compared with vanilla self-attention and previous solutions, our LRSA has a much lower computational cost. 
LRSA can also facilitate attention optimization due to the much shorter token length.
To fit the size of the original $F_{in}$, we then perform a bilinear interpolation after the self-attention calculation.

\para{Complexity and characteristics.}
The computational complexity of LRSA is much lower than existing  self-attention mechanisms for vision transformers.
We summarize the main characteristics and computational complexity of recent popular self-attention mechanisms and our LRSA in \tabref{tab:cmp_attention_complexity}.
Spatial correlation means that self-attention is carried out in the spatial dimension, and some factorized transformers \cite{xu2021coat} compute self-attention in the channel dimension for reducing complexity. 
As can be observed from \tabref{tab:cmp_attention_complexity}, 
other methods often face trade-offs among complexity, 
global receptive field, and spatial correlation. 
In contrast, our LRSA offers advantages in all these aspects.

Let us continue by analyzing the computational complexity of LRSA. For convenience, we do not include the 1D$\leftrightarrow$2D feature reshaping.
LRSA first downsamples the input features $F_{in}\in \mathbb{R}^{N\times C}$ to a fixed size $m \times C$ with a 2D pooling operation, whose computational cost is $O(NC)$. Then, LRSA performs linear transformations and self-attention on the pooled features, which costs $O(mC^2)$.
The computation of self-attention costs $O(m^2C)$.
The final upsampling operation has the same computational cost as downsampling. Overall, the computational complexity of LRSA is $O(NC+mC^2+m^2C)$.
As $m$ is a constant number (\eg, $16^2$) regardless of the value of $N$, we can simplify the complexity of LRSA to $O(NC+C^2)$, which is much smaller than existing methods.

\begin{figure*}[!t]
  \centering
  \includegraphics[width=.9\linewidth]{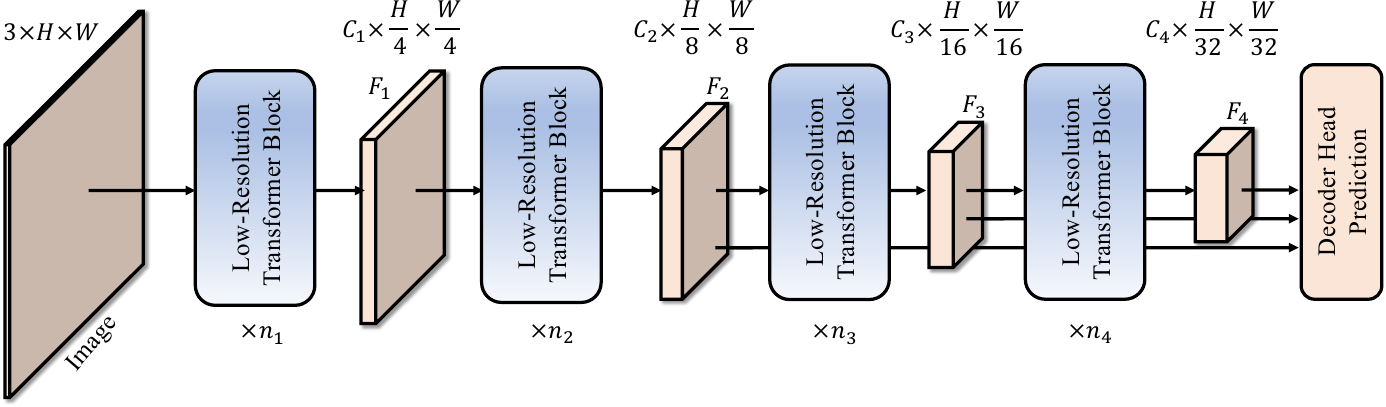}
  \caption{\textbf{Pipeline of the proposed \ourmethod. }
  $F_2$, $F_3$ and $F_4$ are fed into the decoder head for semantic segmentation.
  }
  \label{fig:pipeline}
\end{figure*}

\subsection{Low-Resolution Transformer} \label{sec:lrformer}
In this part, we build the \ourmethod~for semantic segmentation by incorporating the proposed LRSA. The overall architecture of LRFormer is illustrated in \figref{fig:pipeline}, with an encoder-decoder architecture. 

\para{Encoder-decoder.}
Taking a natural image as input, the encoder first downsamples it by a factor of 1/4, following prevailing literature in this field \cite{wang2021pyramid,wang2021pvtv2,liu2021swin,wu2022p2t,fan2021multiscale}.
The encoder consists of four stages with a pyramid structure, each comprising multiple stacked basic blocks.
In between every two stages, we include a patch embedding operation to reduce the feature size by half.
This results in the extraction of multi-level features $F_1, F_2, F_3, F_4$ with strides of 4, 8, 16, and 32, respectively.
We resize $F_2, F_3, F_4$ to the same size as $F_2$ before concatenating and squeezing them to smaller channels.
The resulting features are then fed into our decoder head, which performs further semantic reasoning and outputs the final segmentation map via a $1\times 1$ convolution layer. 
The details of our decoder head are presented in \secref{sec:decoder_head}.

\para{Basic block.}
The basic block is illustrated in \figref{fig:basic_block}.
Like previous transformer blocks \cite{wang2021pyramid,liu2021swin}, the basic block of our LRFormer is composed of a self-attention module and an FFN. 
The FFN is generally an MLP layer composed of two linear layers with GELU \cite{hendrycks2020gaussian} activation in between.
Differently, we renovate the self-attention module with our proposed LRSA.
As LRSA is computed in a very low-resolution space, attaining a low complexity regardless of the input resolution.
However, the low-resolution space may lose the spatial locality of the input features.
Inspired by recent works \cite{chu2021conditional, xie2021segformer, wu2022p2t}, we further introduce depth-wise convolution (DWConv) in both positional encoding and FFN, assisting the feature extraction via capturing spatial local details.
That is, we insert a $3\times 3$ DWConv layer with short connection followed by our LRSA, providing conditional positional encoding \cite{chu2021conditional}.
This strategy is also applied between the two linear layers of the FFN.
Therefore, our basic block can be simply formulated as below:
\begin{equation}
    \label{eq:basic_block}
    \begin{split}
        F_{in}' &= F_{in} + \text{DWConv}(F_{in}),
        \\
        F_{att} &= F_{in}' + \text{LRSA}(\text{LayerNorm}(F_{in}')),
        \\
        F_{out} &= F_{att} + \text{FFN}(\text{LayerNorm}(F_{att})),
    \end{split}
\end{equation}
where $F_{in}$, $F_{att}$ and $F_{out}$ represent the input, output of LRSA, and output of the basic block, respectively.
Since the complexity of DWCons is $O(NC)$, 
inserting DWConvs will still keep the overall self-attention complexity at $O(NC+C^2)$.

\newcommand{\splitcell}[1]{\begin{tabular}{@{}c@{}}#1\end{tabular}}
\newcommand{\bsplitcell}[1]{$\left[\splitcell{#1}\right]$}

\begin{table*}[!tb]
    \centering
    \renewcommand{\arraystretch}{1.2}
    \renewcommand{\tabcolsep}{3.mm}
    \caption{\textbf{Detailed settings of the encoders for different \ourmethod~variants, \ie, T/S/B/L/XL}.
    $C$, $C_h$, $E$, and $n_i$ denote the number of feature channels, channels of each attention head, expansion ratio of FFN, and the number of basic blocks for the $i$-th stage, respectively.
    }
    \label{tab:hyper_param_setting}
    \begin{tabular}{c|c|c|c|c|c|c} \Xhline{0.4mm}
        Stage & Output Size  & \ourmethod-T & \ourmethod-S & \ourmethod-B & \ourmethod-L & \ourmethod-XL
        \\ \hline
        1 & $F_1: \frac{H}{4}\times \frac{W}{4}$ %
        & \splitcell{$C=48$, $E=8$\\ $C_h=24$, $n_1=2$} 
        & \splitcell{$C=64$, $E=8$\\ $C_h=32$, $n_1=3$} 
        & \splitcell{$C=80$, $E=8$\\ $C_h=40$, $n_1=4$} 
        & \splitcell{$C=96$, $E=8$\\ $C_h=48$, $n_1=4$}
        & \splitcell{$C=128$, $E=8$\\ $C_h=64$, $n_1=4$}
        \\ \hline
        2 & $F_2: \frac{H}{8}\times \frac{W}{8}$ %
        & \splitcell{$C=96$, $E=8$\\ $C_h=24$, $n_2=2$} 
        & \splitcell{$C=128$, $E=8$\\ $C_h=32$, $n_2=3$} 
        & \splitcell{$C=160$, $E=8$\\ $C_h=40$, $n_2=4$} 
        & \splitcell{$C=192$, $E=8$\\ $C_h=48$, $n_2=6$} 
        & \splitcell{$C=256$, $E=8$\\ $C_h=64$, $n_1=8$}
        \\ \hline
        3 & $F_3: \frac{H}{16}\times \frac{W}{16}$ %
        & \splitcell{$C=240$, $E=4$\\ $C_h=24$, $n_3=6$} 
        & \splitcell{$C=320$, $E=4$\\ $C_h=32$, $n_3=12$}
        & \splitcell{$C=400$, $E=4$\\ $C_h=40$, $n_3=15$}
        & \splitcell{$C=480$, $E=4$\\ $C_h=48$, $n_3=18$}
        & \splitcell{$C=640$, $E=4$\\ $C_h=64$, $n_3=22$}
        \\ \hline
        4 & $F_4: \frac{H}{32}\times \frac{W}{32}$ %
        & \splitcell{$C=384$, $E=4$ \\ $C_h=24$, $n_4=3$} %
        & \splitcell{$C=512$, $E=4$ \\ $C_h=32$, $n_4=3$} 
        & \splitcell{$C=512$, $E=4$ \\ $C_h=32$, $n_4=8$} 
        & \splitcell{$C=640$, $E=4$ \\ $C_h=40$, $n_4=8$} 
        & \splitcell{$C=768$, $E=4$ \\ $C_h=48$, $n_4=8$} 
        \\ 
        \Xhline{0.4mm}
    \end{tabular}
\end{table*}

\para{Architecture setting.}
To fit the budgets of different computational resources, we design four variants of \ourmethod, namely LRFormer-T/S/B/L, stacking different numbers of basic blocks for each stage in the encoder. We summarize the detailed settings of their encoders in \tabref{tab:hyper_param_setting}.
In terms of ImageNet pretraining \cite{russakovsky2015imagenet}, the computational cost of LRFormer-T/S/B/L is similar to ResNet-18 \cite{he2016deep} and Swin-T/S/B \cite{liu2021swin}, respectively.

\subsection{Decoder Head} \label{sec:decoder_head}
In semantic segmentation, it is suboptimal to predict the results based solely on the final output of the encoder, as multi-level information is useful in perceiving objects with various scales and aspect ratios \cite{cheng2021mask2former,xie2021segformer}. Thus, we design a simple decoder for LRFormer to aggregate multi-level features efficiently and effectively.
To this end, we note that an MLP aggregation can achieve good performance in the state-of-the-art work SegFormer \cite{xie2021segformer}. However, it does not consider the spatial correlation between the features from different levels.
Therefore, we encapsulate our LRSA into our decoder for feature refinement, strengthening the semantic reasoning of LRFormer.

As mentioned above, $F_2, F_3, F_4$ are resized to the same size as $F_2$ and then concatenated together. We apply a $1\times 1$ convolution on the concatenated feature to squeeze the number of channels. Then, a basic block (LRSA + FFN) is adopted to refine the squeezed feature. As we know, the feature from the top of the encoder, \ie, $F_4$, could be the most semantic meaningful. To avoid the loss of semantic information in the aggregation of high-level ($F_4$) and low-level ($F_2, F_3$) features, we concatenate the refined feature with $F_4$ to enhance the semantics. After that, another basic block is connected for further feature refinement. Finally, we infer the segmentation prediction from the refined feature with a simple $1\times 1$ convolution. The experiments demonstrate that our simple decoder with LRSA can do better than previous state-of-the-art decoder heads for semantic segmentation, as shown in \tabref{tab:exp_decoder_cmp}.

\subsection{Implementation Details}
\label{sec:implement_details}

In LRFormer, we apply the overlapped patch embedding, \ie, a $3\times 3$ convolution with a stride of 2, to downsample the features by half between each stage.
To strengthen multi-scale learning of LRSA with negligible cost, we use pyramid pooling \cite{wu2022p2t} to extract multi-scale features when computing the key and value features in LRSA.
The desired fixed downsampling size $m$ for generating the query, key and value is $16^2$ for semantic segmentation. Such size is changed to $7^2$ for ImageNet pretraining because $m=16^2$ is too large for image classification.
For the number of channels in the decoder, we set it to 256/384/512/640 for LRFormer-T/S/B/L, respectively.

\section{Experiments}

\subsection{Experimental Setup}
\label{sec:setup}

\para{Datasets.}
We perform experiments on three well-established datasets.
ADE20K \cite{zhou2017scene} is a very challenging scene parsing dataset that contains 150 semantic classes with diverse foreground and background, consisting of 20K, 2K, and 3.3K images for training, validation, and testing, respectively.
COCO-Stuff \cite{caesar2018coco} labels both things and stuffs with a total of 171 fine-grained semantic labels, with 164K, 5K, 20K, and 20K images for training, validation, test-dev, and test challenge.
Cityscapes \cite{cordts2016cityscapes} is a high-quality dataset for street scene parsing that contains 3K, 0.5K, and 1.5K driving images for training, validation, and testing. 
These datasets cover a wide range of semantic categories and pose different challenges for semantic segmentation models.

\para{ImageNet pretraining.}
We adopt the popular \textit{timm} package to implement our network.
Following other networks, we first pretrain the backbone encoder of LRFormer on the ImageNet-1K dataset, which has 1.3M training and 50K validation images with 1K object categories.
During ImageNet pretraining, the decoder head of LRFormer is omitted.
To regularize the training process, we follow the standard data augmentation techniques and optimization strategy used in previous works \cite{touvron2021training,xie2021segformer,liu2021swin}.
We use AdamW \cite{loshchilov2017decoupled} as the default optimizer with a learning rate of 0.001, weight decay of 0.05, a \textit{cosine} learning rate adjustment schedule, and a batch size of 1024. 
No model EMA is applied. 
The backbone encoder is pretrained for 300 epochs, and we apply layer scale \cite{touvron2021going} to alleviate the overfitting of large networks, as suggested by recent works \cite{liu2022convnet, touvron2021going}. 
For \ourmethod-L, we follow \cite{liu2021swin, liu2022convnet} additionally pretrain the network on the full ImageNet-22K dataset for 90 epochs and then finetune it on ImageNet-1K dataset for 30 epochs. 
In the finetuning, the learning rate is set as 5e-5, and each mini-batch has 512 images.

\begin{table}[!t]
  \centering
  \setlength{\tabcolsep}{2mm}
  \caption{\textbf{Comparisons with recent methods on the ADE20K dataset \cite{zhou2017scene}.}
  The results of our method are marked as \textbf{bold}.
  ``$^\dagger$'' indicates the result pretrained on ImageNet-22K.
  }
  \label{tab:exp_ade20k}
\resizebox{\columnwidth}{!}{%
  \begin{tabular}{l|c|c|c} \Xhline{1pt}
      Method
       & FLOPs $\downarrow$ & \#Params $\downarrow$ & mIoU $\uparrow$
      \\ \Xhline{1pt}
      SegFormer-B1 \cite{xie2021segformer}  & 16G & 14M & 42.2\%  \\
      Vim-Ti \cite{zhu2024vision} & - & 13M & 41.0\% \\
      HRFormer-S \cite{yuan2021hrformer}  & 109G & 14M & 44.0\%\\
      \ourmethod-T (Ours) & 17G & 13M  & \textbf{46.7\%} \\
      \hline
      SegFormer-B2 \cite{xie2021segformer}  & 62G & 28M & 46.5\% \\
      P2T-Small \cite{wu2022p2t} & 43G & 28M & 46.7\% \\
      MaskFormer \cite{cheng2021per} & 55G & 42M  & 46.7\% \\
      FeedFormer-B2 \cite{shim2023feedformer} & 43G & 29M & 48.0\% \\
      Mask2Former \cite{cheng2021mask2former} & 74G & 47M  & 47.7\% \\
      \ourmethod-S (Ours)  & 40G & 32M & \textbf{50.0\%} \\
      \hline
      HRFormer-B \cite{yuan2021hrformer}  & 280G & 56M & 48.7\% \\
      Vim-S \cite{zhu2024vision} & - & 46M & 44.9\% \\
      SegFormer-B3 \cite{xie2021segformer}  & 96G & 47M & 49.4\% \\
      \ourmethod-B (Ours) & 75G & 69M  & \textbf{51.0\%} \\
      \hline
      DPT-Hybrid \cite{ranftl2021vision} & 308G  & 124M & 49.0\% \\
      SegFormer-B5 \cite{xie2021segformer} & 183G & 85M  & 51.0\% \\
      DAViT-B \cite{ding2022davit}  & 294G & 121M & 49.4\% \\
      FasterViT-4 \cite{hatamizadeh2023fastervit} & 323G & 457M & 49.1\% \\
      InternImage-B \cite{wang2023internimage} & 296G & 128M & 50.8\% \\
      MaskFormer \cite{cheng2021per}  & 195G & 102M & 51.3\% \\

      \ourmethod-L (Ours)  & 183G & 113M & \textbf{52.6\%} \\
      \hline
      SETR-MLA$^\dagger$ \cite{zheng2021rethinking}  & - & 302M & 48.6\% \\
      MaskFormer$^\dagger$ \cite{cheng2021mask2former}  & 195G & 102M & 53.1\% \\
      CSWin-B$^\dagger$ \cite{dong2022cswin} & 463G & 109M  & 51.8\%  \\
      \ourmethod-L$^\dagger$ (Ours)  & 183G & 113M & \textbf{54.2\%} \\
      \Xhline{1pt}
  \end{tabular}}
\end{table}

\para{Training for semantic segmentation.}
We use \textit{mmsegmentation} framework to train our network for semantic segmentation.
AdamW \cite{loshchilov2017decoupled} is adopted as the default optimizer, with learning of 0.00006, weight decay of 0.01, and \textit{poly} learning rate schedule with factor 1.0.
Following \cite{liu2021swin, xie2021segformer}, the weight decay of LayerNorm \cite{ba2016layer} layers is set as 0.
Regarding the data augmentation, we use the same strategy as mentioned in \cite{liu2021swin, xie2021segformer}.
That is we construct the pipeline of image resizing ($0.5\sim 2\times$), 
random horizontal flipping, 
followed by a random cropping of size 512$\times$512, 512$\times$512, and 1024$\times$1024 for ADE20K, COCO-Stuff, and Cityscapes datasets, respectively.
Note that for our largest model \ourmethod-L in ADE20K, the cropped size remains 640$\times$640, consistent with recent works. 
The mini-batch size is set to 16, 16, and 8 images for ADE20K, COCO-Stuff, and Cityscapes datasets, respectively. We train our network for 160K, 80K, and 160K iterations for ADE20K, COCO-Stuff, and Cityscapes datasets, respectively. 
We only use the cross-entropy loss for training and do not employ any extra losses like the auxiliary loss \cite{zhao2017pyramid} and OHEM \cite{shrivastava2016training}.

\para{Testing for semantic segmentation.}
During testing, we maintain the original aspect ratio of the input image and resize it to a shorter size of 512 and a longer size not exceeding 2048 for the ADE20K and COCO-Stuff datasets.
We follow the suggestion of \cite{xie2021segformer} and resize the input size of LRFormer-L for the ADE20K dataset to a shorter size of 640 and a longer size not exceeding 2560.
In the Cityscapes dataset, we apply a crop size of 1024$\times$1024 with sliding window testing strategy following \cite{xie2021segformer}.

\subsection{Comparisons}

\para{ADE20K.}
Results are shown in \tabref{tab:exp_ade20k}.
LRFormer is compared with several recent transformer-based and Mamba-based methods in different complexity levels. Results of other methods are from their official repositories.
We can observe that our LRFormer exhibits strong superiority over other methods.
In terms of the mIoU, LRFormer-T/S/B/L are 4.5\%/3.5\%/2.6\%/1.6\% better than SegFormer-B1/B2/B4/B5 \cite{liu2021swin, xie2021segformer}.
LRFormer-T is 2.3\% better than Swin-T-based Mask2Former \cite{cheng2021mask2former} with near half FLOPs.
With ImageNet-22K pretraining, LRFormer is 1.1\% and 2.4\% better than the strongest Swin-B-based MaskFormer \cite{liu2021swin, cheng2021per} and UperNet-based CSwin \cite{xiao2018unified,dong2022cswin} with fewer FLOPs.
Compared with the representative Mamba-based Vim \cite{zhu2024vision} with linear complexity, our LRFormer is significantly better.
The visualized \figref{fig:cmp_ade20k} of accuracy-FLOPs shows a more intuitive view of the comparisons.

\begin{table}[!t]
  \centering
  \setlength{\tabcolsep}{2mm}
  \caption{\textbf{Comparisons with recent transformer-based methods on the full COCO-Stuff dataset \cite{caesar2018coco}.}
  Results of our method are marked as \textbf{bold}.
  }
  \label{tab:exp_cocostuff}
\resizebox{\columnwidth}{!}{%
  \begin{tabular}{l|c|c|c} \Xhline{1pt}
      Method
       & FLOPs $\downarrow$ & \#Params $\downarrow$ & mIoU $\uparrow$
      \\ \Xhline{1pt}
      HRFormer-S \cite{yuan2021hrformer}  & 109G & 14M & 37.9\% \\
      SegFormer-B1 \cite{xie2021segformer}  & 16G & 14M & 40.2\% \\
      \ourmethod-T (Ours) & 17G & 13M & \textbf{43.9\%} \\
      \hline
      SegFormer-B2 \cite{xie2021segformer}  & 62G & 28M & 44.6\% \\
      \ourmethod-S (Ours) & 40G & 32M  & \textbf{46.4\%} \\
      \hline
      HRFormer-B \cite{yuan2021hrformer}  & 280G & 56M & 42.4\% \\
      SegFormer-B3 \cite{xie2021segformer}  & 79G & 47M & 45.5\% \\
      SegFormer-B5 \cite{xie2021segformer}  & 112G & 85M & 46.7\% \\
      \ourmethod-B (Ours)  & 75G & 69M & \textbf{47.2\%} \\
      \ourmethod-L (Ours) & 122G  & 113M & \textbf{47.9\%} \\
      \Xhline{1pt}
  \end{tabular}}
\end{table}

\begin{table}[!t]
  \centering
  \setlength{\tabcolsep}{2mm}
  \caption{\textbf{Comparisons with recent transformer-based methods on the Cityscapes dataset \cite{cordts2016cityscapes}.}
  Results of our method are marked as \textbf{bold}.
  FLOPs are calculated for an input size of $1024\times 2048$.
  }
\resizebox{\columnwidth}{!}{%
  \begin{tabular}{l|c|c|c} \Xhline{1pt}
      Method
       & FLOPs $\downarrow$ & \#Params $\downarrow$ & mIoU $\uparrow$
      \\ \Xhline{1pt}
      HRFormer-S \cite{yuan2021hrformer} & 872G & 14M & 80.0\%\\
      SegFormer-B1 \cite{xie2021segformer} & 244G & 14M  & 78.5\% \\
      \ourmethod-T (Ours)  & 122G & 13M & \textbf{80.7\%} \\
      \hline
      SegFormer-B2 \cite{xie2021segformer}  & 717G  & 28M & 81.0\% \\
      \ourmethod-S (Ours) & 295G & 32M  & \textbf{81.9\%} \\
      \hline
      HRFormer-B \cite{yuan2021hrformer} & 2240G & 56M  & 81.9\% \\
      SegFormer-B3 \cite{xie2021segformer} & 963G  & 47M & 81.7\% \\
      SegFormer-B5 \cite{xie2021segformer} & 1460G & 85M  & 82.4\% \\
      \ourmethod-B (Ours) & 555G & 67M  & \textbf{83.0\%} \\
      \ourmethod-L (Ours)  & 908G & 111M & \textbf{83.2\%} \\
      \Xhline{1pt}
  \end{tabular}}
  \label{tab:exp_cityscapes}
\end{table}

\para{COCO-Stuff.}
We elaborate the results in \tabref{tab:exp_cocostuff}.
We evaluated our method on different network scales and compared it against recent popular methods. LRFormer achieved the highest mIoU on all network scales, outperforming the other methods. Specifically, our \ourmethod-T model achieved a mIoU of 43.9\%, which is 3.7\% higher than HRFormer-S and 3.7\% higher than SegFormer-B1. Similarly, our \ourmethod-S and \ourmethod-B models outperformed the corresponding SegFormer models by 1.8\% and 1.7\%. Our \ourmethod-L model outperforms SegFormer-B5 by 1.2\%. 
These experimental comparisons demonstrate the superiority of \ourmethod~on the COCO-Stuff dataset.

\para{Cityscapes.}
\tabref{tab:exp_cityscapes} presents the experimental comparisons between LRFormer and recent popular methods on the Cityscapes dataset.
LRFormer outperforms SegFormer and HRFormer in all cases. 
We can observe that due to large input size, FLOPs of other methods are much higher than ours. 
For example, SegFormer-B2 costs 717G FLOPs while our LRFormer-S only spends 41\% FLOPs with 0.9\% improvement.
More complexity analysis can refer to \tabref{tab:memory_flops}.

\begin{table}[t]
  \setlength\tabcolsep{1pt}
  \small
  \renewcommand{\arraystretch}{1.}
  \centering
  \caption{\textbf{Classification results on ImageNet-1K \cite{russakovsky2015imagenet} dataset}.
  Results of our method are marked as \textbf{bold}.
  Results marked with ``$^\dagger$'' are pretrained on ImageNet-22K dataset. 
  }\label{tab:in1k}
  \begin{tabular}{lccccc} 
  \Xhline{1pt}
  Model  & FLOPs  $\downarrow$ & \#Params $\downarrow$ & Size & Top-1 Acc. $\uparrow$\\ 
  \Xhline{1pt}
  PVTv2-B1~\cite{wang2021pvtv2}  & 2.1G & 13M & $224^2$ & 78.7\% \\
  HAT-Net-T~\cite{liu2021vision} & 2.0G & 13M & $224^2$ & 79.8\% \\
  P2T-Tiny~\cite{wu2022p2t}  & 1.8G & 12M & $224^2$ & 79.8\% \\
  \ourmethod-T (Ours) & 1.8G & 13M & $224^2$ & \textbf{80.8\%} \\
  \hline
  Swin-T~\cite{liu2021swin}  & 4.5G & 28M & $224^2$ & 81.5\% \\
  MViTv2-T \cite{li2022mvitv2}  & 4.7G & 24M & $224^2$ & 82.3\% \\
  Vim-S \cite{zhu2024vision} & - & 26M & $224^2$ &  81.4\% \\
  HAT-Net-S~\cite{liu2021vision} & 4.3G & 26M & $224^2$ & 82.6\% \\
  ConvNeXt-T~\cite{liu2022convnet}  & 4.5G & 29M & $224^2$ & 82.1\% \\
  \ourmethod-S (Ours) & 4.7G & 30M & $224^2$ & \textbf{83.5\%} \\
  \hline
  Swin-S~\cite{liu2021swin}  & 8.7G & 50M & $224^2$ & 83.0\% \\
  ConvNeXt-S~\cite{liu2022convnet} & 8.7G & 50M & $224^2$ & 83.1\% \\
  DAT-S \cite{xia2022vision} & 9.0G & 50M & $224^2$ & 83.7\% \\
  P2T-Large \cite{wu2022p2t} & 9.8G & 55M & $224^2$ & 83.9\% \\
  \ourmethod-B (Ours) & 9.3G & 62M & $224^2$ & \textbf{84.5\%} \\
  \hline
  DeiT-B~\cite{touvron2020training} & 17.5G & 86M  & $224^2$ & 81.8\% \\
  RegNetY-16G~\cite{regnet}  & 16.0G & 84M & $224^2$ & 82.9\% \\
  RepLKNet-31B~\cite{ding2022scaling}  & 15.3G  & 79M & $224^2$ & 83.5\% \\
  SwinT-B~\cite{liu2021swin} & 15.4G & 88M  & $224^2$ & 83.5\% \\
  ConvNeXt-B~\cite{liu2022convnet}& 15.4G  & 89M & $224^2$ & 83.8\% \\
  FocalNet-B~\cite{yang2022focal} & 15.4G & 89M & $224^2$ & 83.9\% \\
  CSwin-B~\cite{dong2022cswin} & 15.0G & 78M  & $224^2$ & 84.2\% \\
  DAT-B \cite{xia2022vision} & 15.8G & 88M  & $224^2$ & 84.0\% \\
  Vim-B~\cite{zhu2024vision} & - & 98M & $224^2$ & 83.2\% \\
  \ourmethod-L (Ours) & 15.7G & 101M  & $224^2$ & \textbf{85.0\%} \\
  \hline
  Swin-B$^\dagger$~\cite{liu2021swin}  & 15.4G & 88M & $224^2$ & 85.2\% \\
  ConvNeXt-B$^\dagger$~\cite{liu2022convnet}  & 15.4G  & 89M & $224^2$ & 85.8\% \\
  \ourmethod-L$^\dagger$ (Ours)  & 15.7G & 101M & $224^2$ & \textbf{86.4\%} \\
  ConvNeXt-B$^\dagger$~\cite{liu2022convnet} & 45.1G & 89M  & $384^2$ & 86.8\% \\
  Swin-B$^\dagger$~\cite{liu2021swin} & 47.0G & 88M  & $384^2$ & 86.4\% \\
  \ourmethod-L$^\dagger$ (Ours) & 46.3G & 101M  & $384^2$ & \textbf{87.2\%} \\
  \hline
  Swin-L$^\dagger$~\cite{liu2021swin} & 34.5G & 197M & $224^2$ & 86.3\% \\
  ConvNeXt-L$^\dagger$~\cite{liu2022convnet} & 34.4G & 198M & $224^2$ & 86.6\% \\
   \ourmethod-XL$^\dagger$ (Ours) & 31.6G & 187M  & $224^2$ & \textbf{87.0\%} \\
  \Xhline{1pt}
  \end{tabular}
\end{table}

\para{ImageNet.}
Since we pretrained our backbone encoder on ImageNet, we also evaluate our network on ImageNet classification only for reference.
Results are shown in \tabref{tab:in1k}.
We divide them to five groups. The four groups are divided by the FLOPs of approximate 2G, 4.5G, 9G, 16G, respectively.
The fifth and sixth groups include results pretrained on ImageNet-22K dataset.
The backbone encoder of our LRFormer outperformed recent state-of-the-art CNN-based methods such as ConvNeXt \cite{liu2022convnet} and RepLKNet \cite{ding2022scaling}, and transformer-based methods like DAT \cite{xia2022vision} and P2T \cite{wu2022p2t}.

\begin{table}[t]
  \centering
  \setlength{\tabcolsep}{1.5mm}
  \caption{\textbf{Experiments on the fixed pooled size settings of our LRSA.}
  The performance is saturated when pooled size is larger than $16\times 16$.
  }
  \label{tab:pool_size}
\resizebox{\columnwidth}{!}{%
\label{tab:exp_pooled_size}
  \begin{tabular}{l|llc} \Xhline{1pt}
     Pooled Size $\downarrow$ 
     & FLOPs $\downarrow$  & Training Memory $\downarrow$ 
      & mIoU $\uparrow$
      \\ \Xhline{1pt}
      $4\times 4$
      & 38G (-5\%) & 3.9GB (-7\%)  & 46.3\%  \\
      $8\times 8$  
      & 38G (-5\%) & 4.0GB (-5\%)  & 46.8\%  \\
      $16\times 16$  
      & 40G & 4.2GB & 48.5\%  \\
      $32\times 32$   
       & 52G (+30\%) & 5.3GB (+26\%) & 48.6\%  \\
       $48\times 48$   
       & 74G (+85\%) & 7.4GB (+76\%)  & 48.7\%  \\
       $64\times 64$   
       & 108G (+170\%) & 10.9GB (+160\%) & 48.5\%  \\
  \Xhline{1pt}
  \end{tabular}}
\end{table}

\subsection{Visualization analysis.}
To visually illustrate the effectiveness of our method, 
we pick segformer\cite{xie2021segformer} as the model for intuitive comparison from ADE20K val set and Cityscapes val set, 
as shown in \figref{fig:vis_ade20k} and \figref{fig:vis_cityscapes} respectively. 
The results indicate that LRFormer is capable of generating more precise segmentation maps, particularly in the areas highlighted by the red boxes.
We discover that LRFormer offers significant advantages in terms of maintaining object segmentation integrity and capturing intricate details.

\newcommand{\AddImg}[1]{\includegraphics[width=.242\linewidth]{#1}}

\begin{figure}[!t]
  \centering

  \setlength\tabcolsep{1pt}
  \begin{tabular}{cccc}
  \AddImg{ADE_val_00001088} & \AddImg{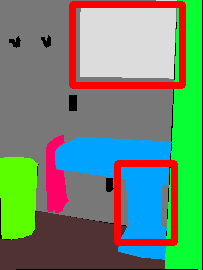} &
  \AddImg{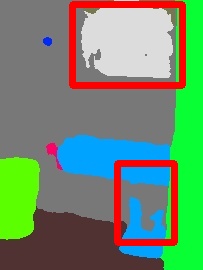} & \AddImg{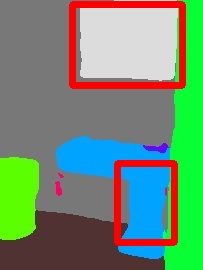}
  \\ 
  \AddImg{ADE_val_00001367} & \AddImg{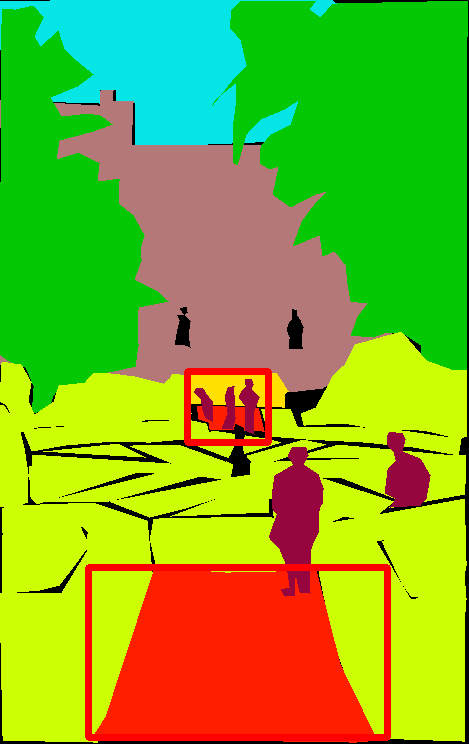}&
  \AddImg{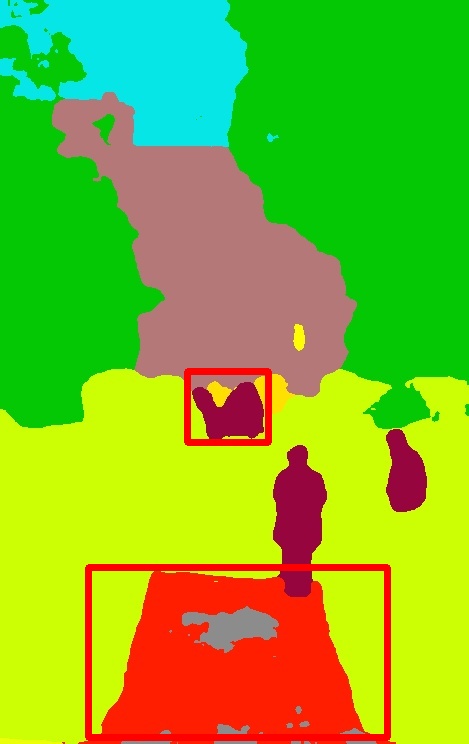} & \AddImg{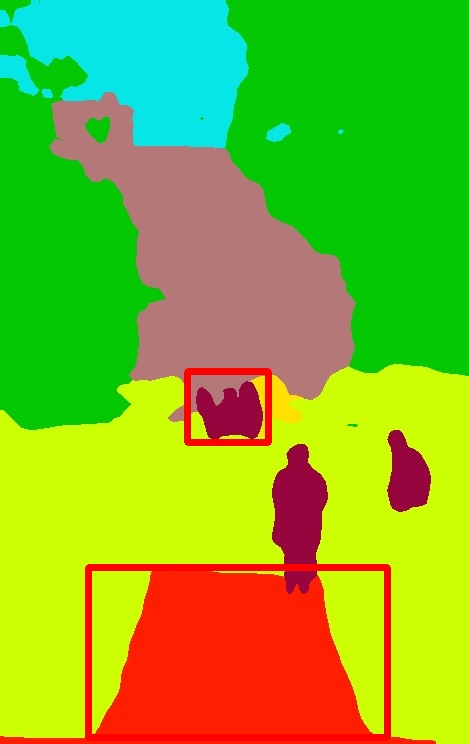}
  \\ 
  \AddImg{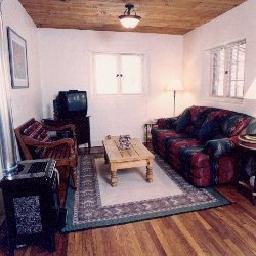} & \AddImg{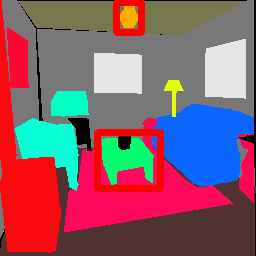} &
  \AddImg{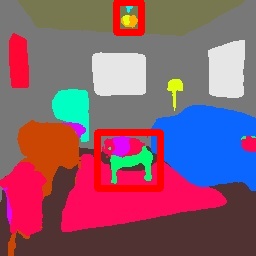} & \AddImg{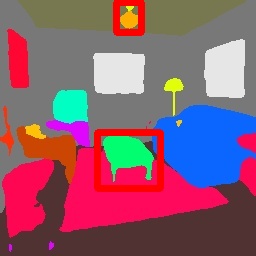}
  \\ 
  \AddImg{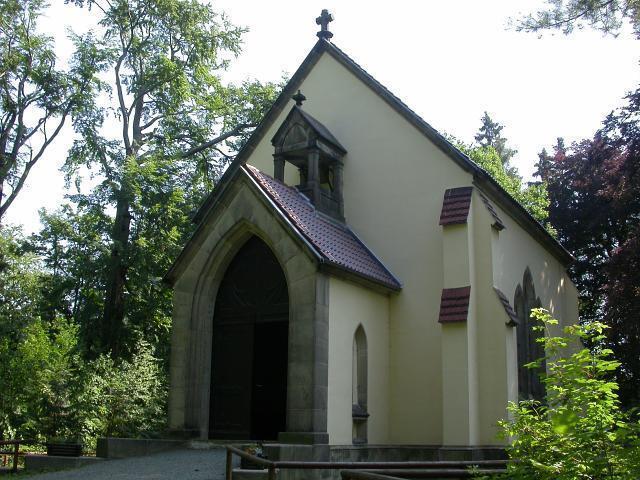} & \AddImg{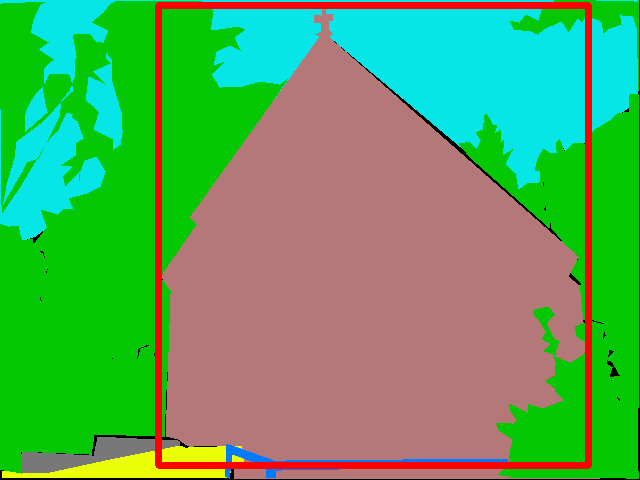} &
  \AddImg{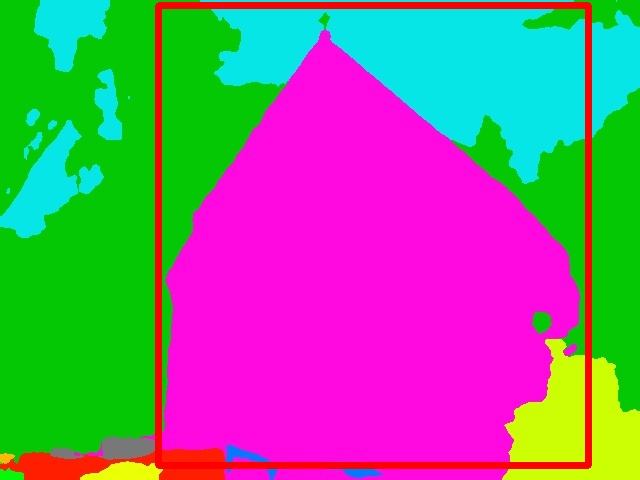} & \AddImg{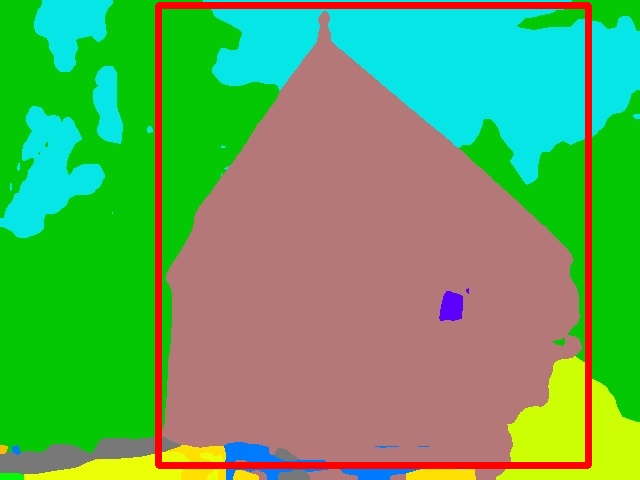}
  \\
  \end{tabular}
  \\
  \caption{\textbf{Qualitative Visualization on ADE20K val set.} The figures from left to right are input images, ground truth, segmentation maps of SegFormer\cite{xie2021segformer}, segmentation maps of our LRFormer. Significant improvements are indicated by red boxes on segmentation maps.}
  \label{fig:vis_ade20k}
\end{figure}

\begin{figure*}[!t]
  \centering
  \AddImg{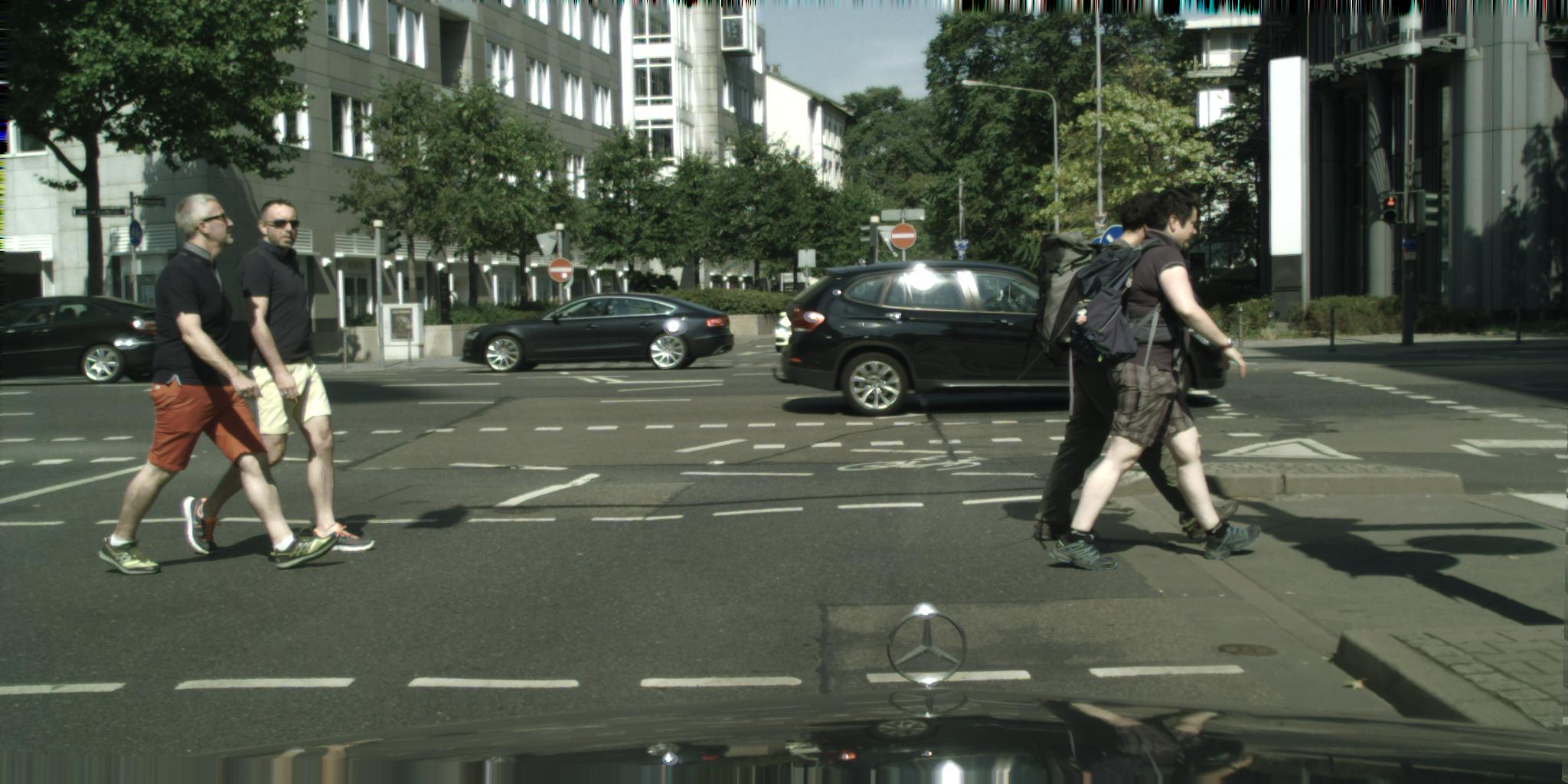} \hfill 
  \AddImg{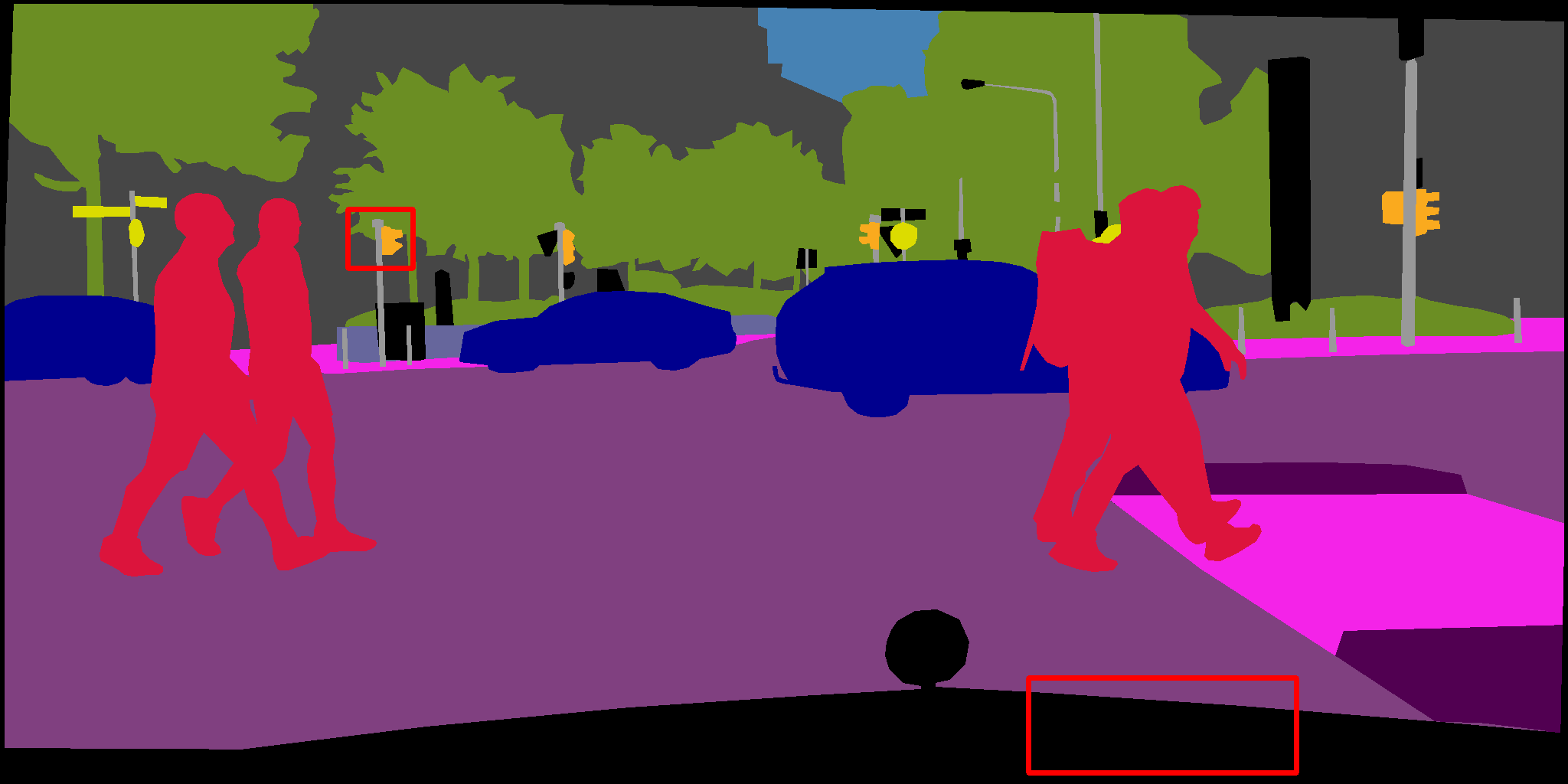} \hfill
  \AddImg{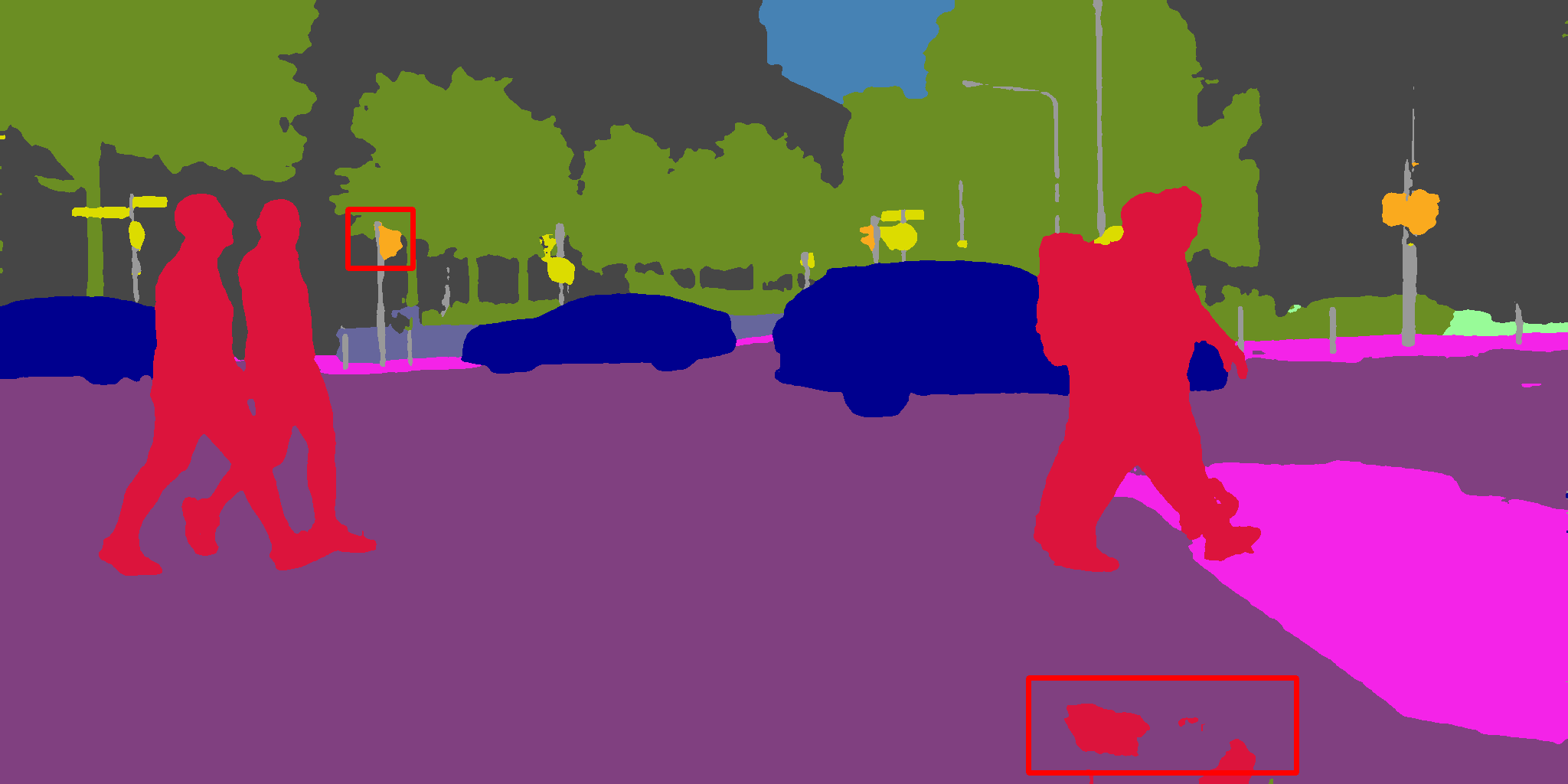} \hfill 
  \AddImg{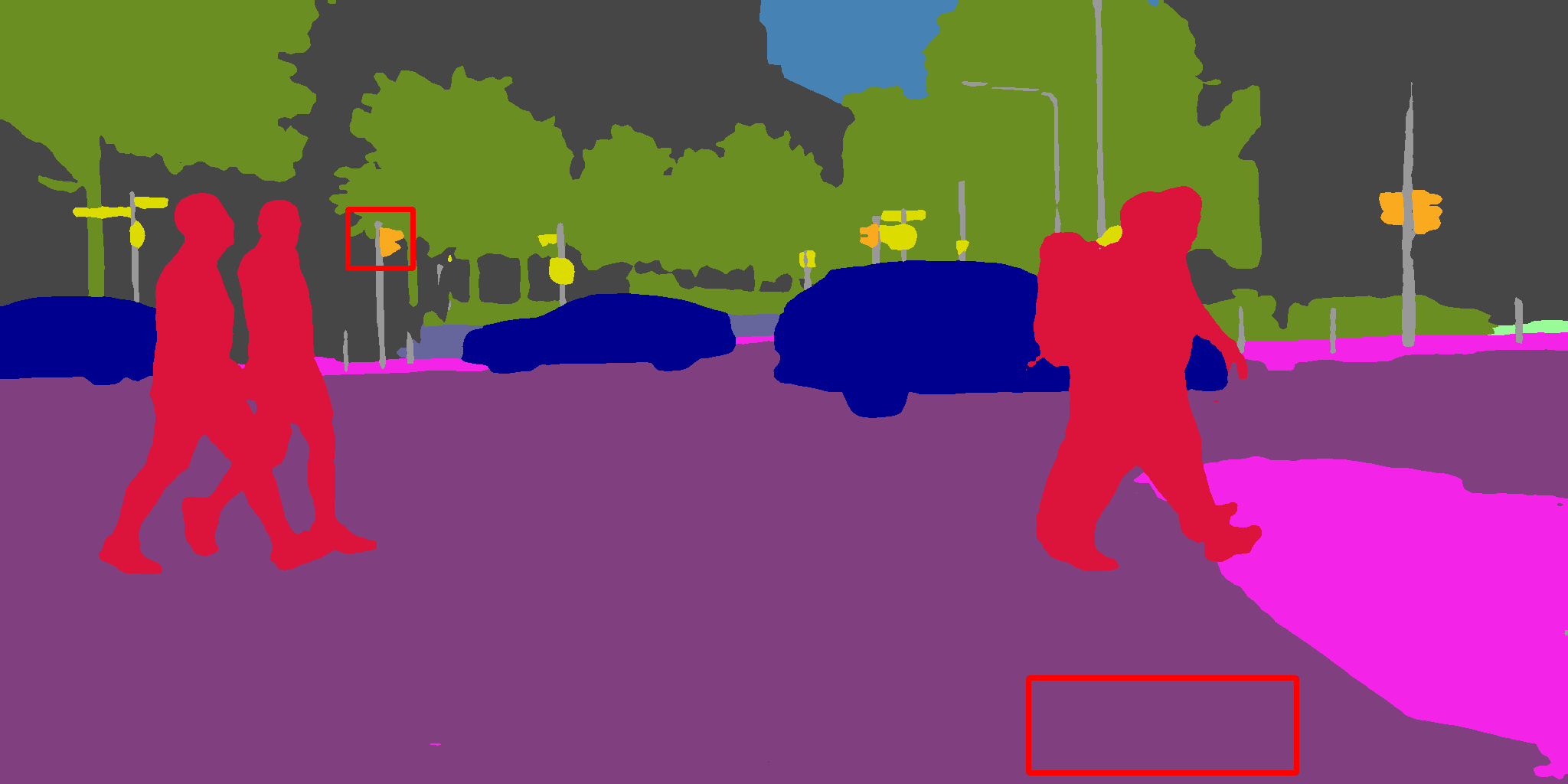}
  \\ \vspace{0.04in}
  \AddImg{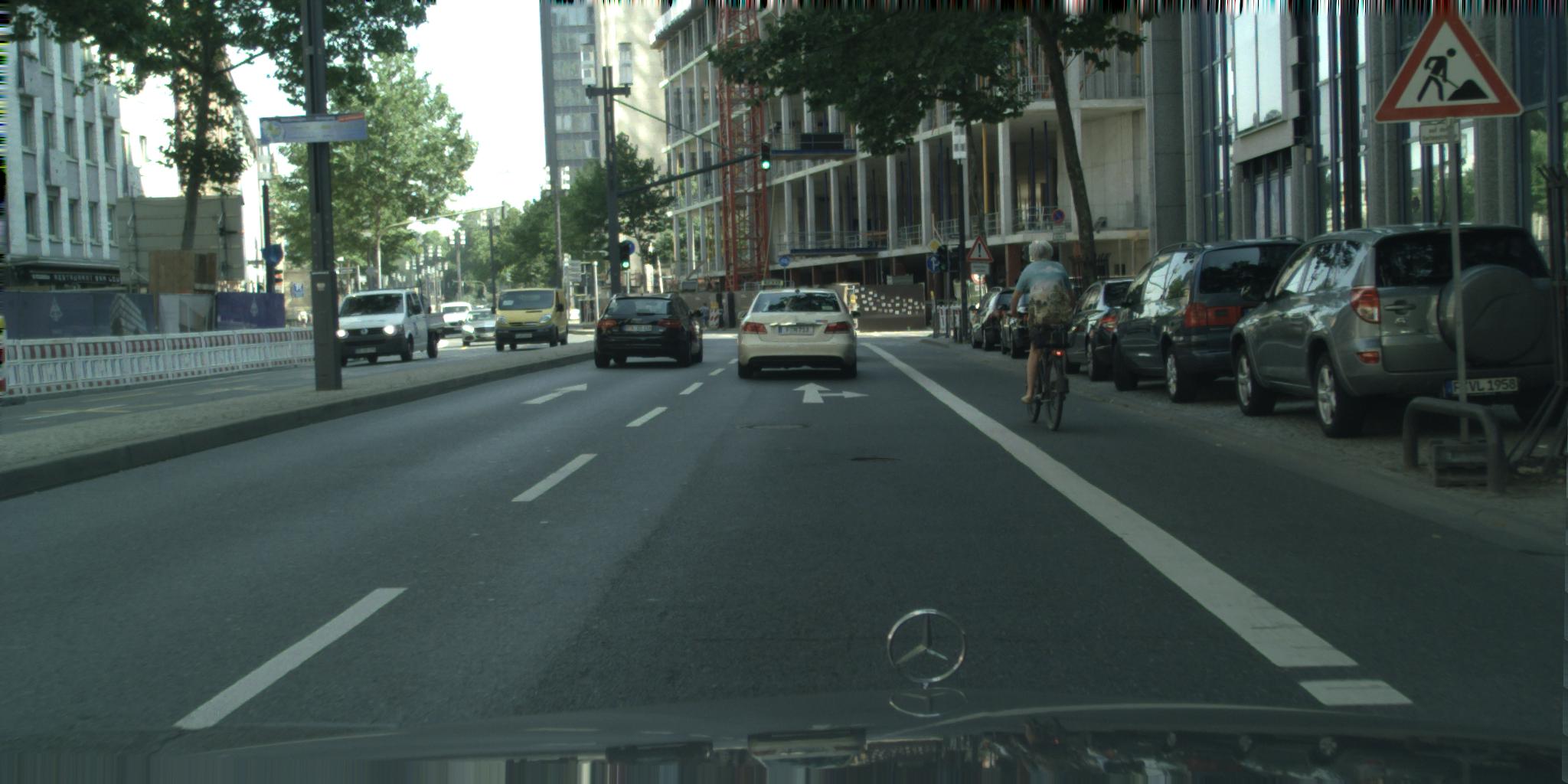} \hfill 
  \AddImg{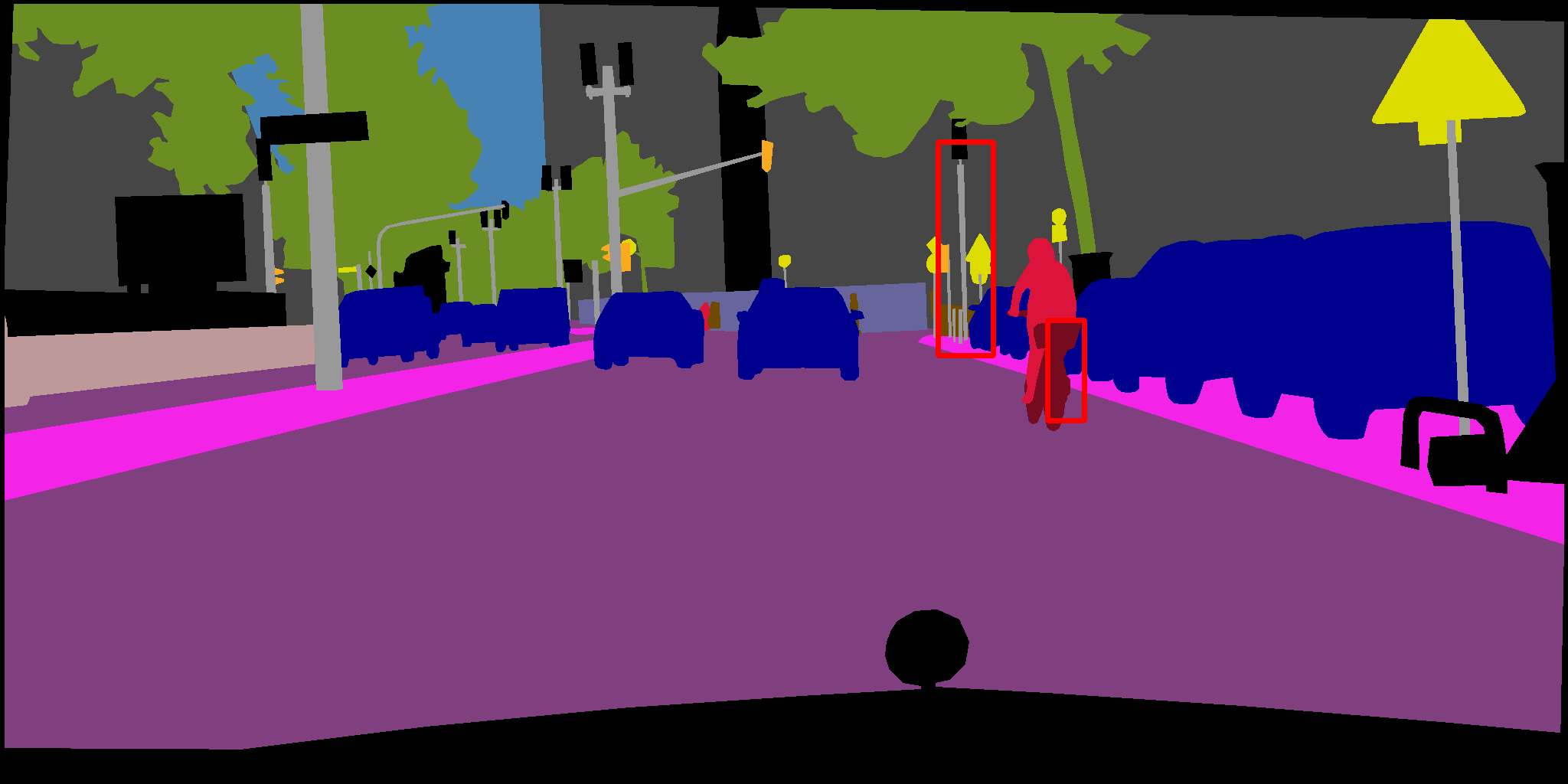} \hfill
  \AddImg{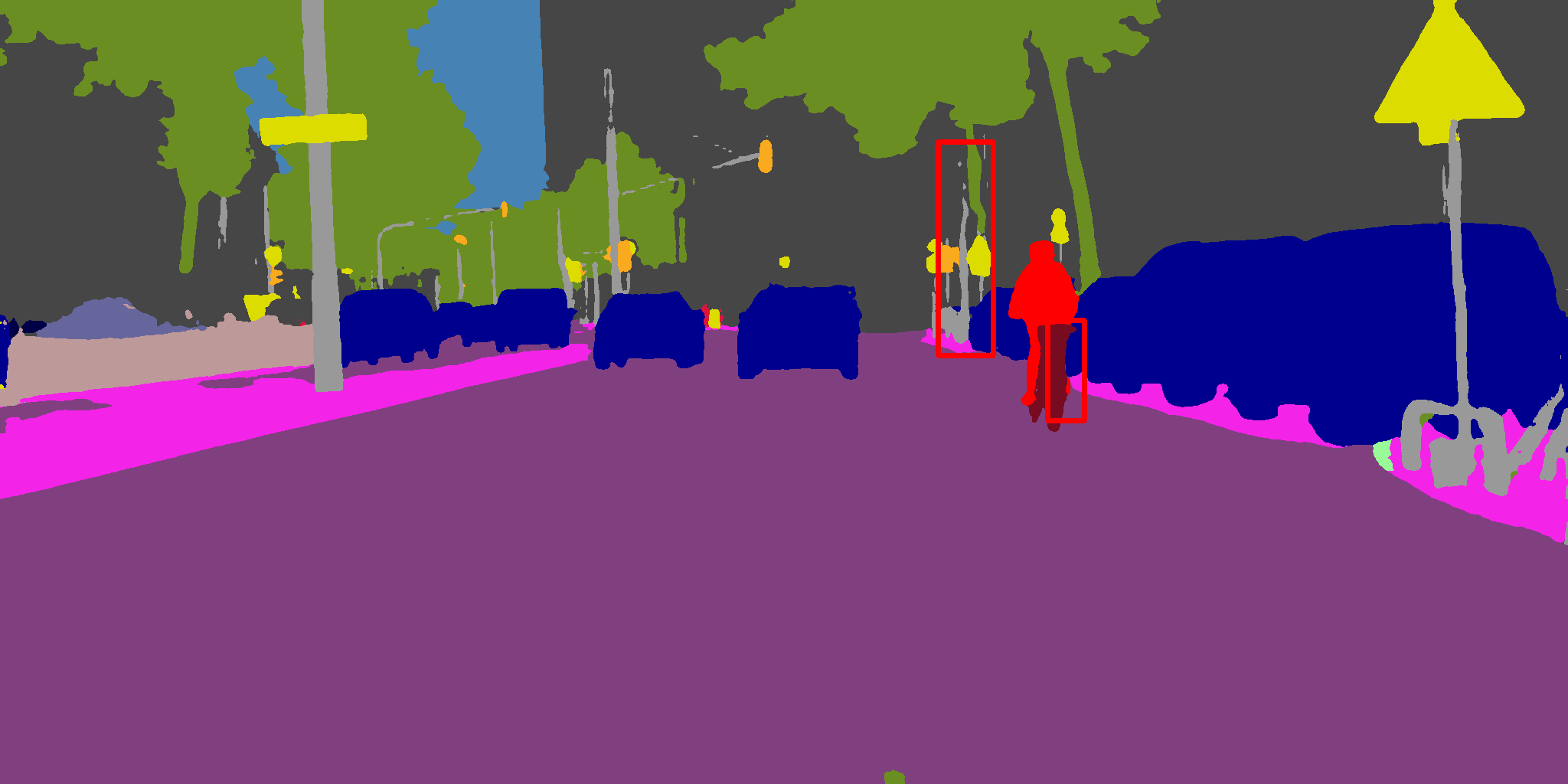} \hfill 
  \AddImg{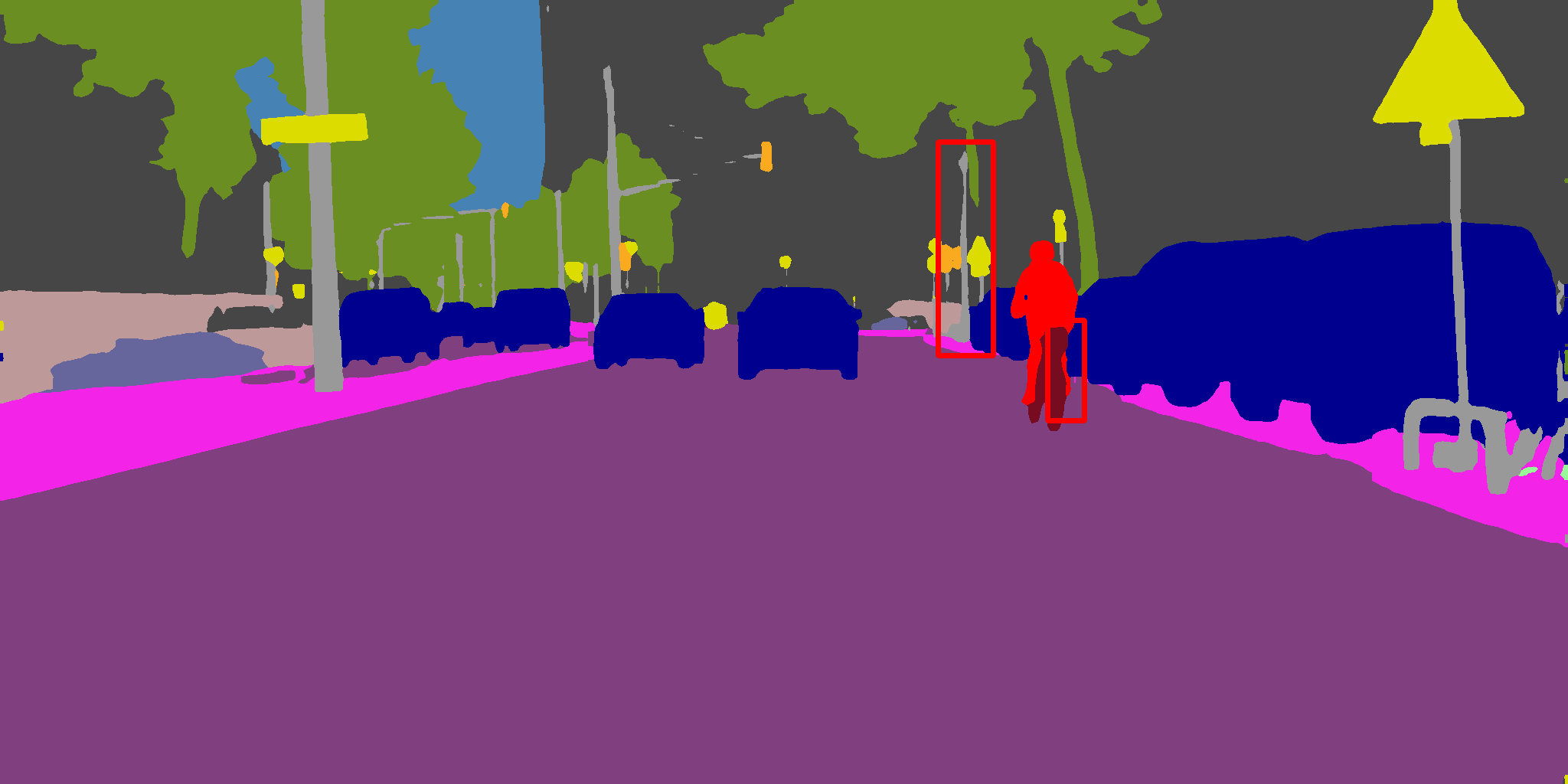}
  \\ \vspace{0.04in}
  \AddImg{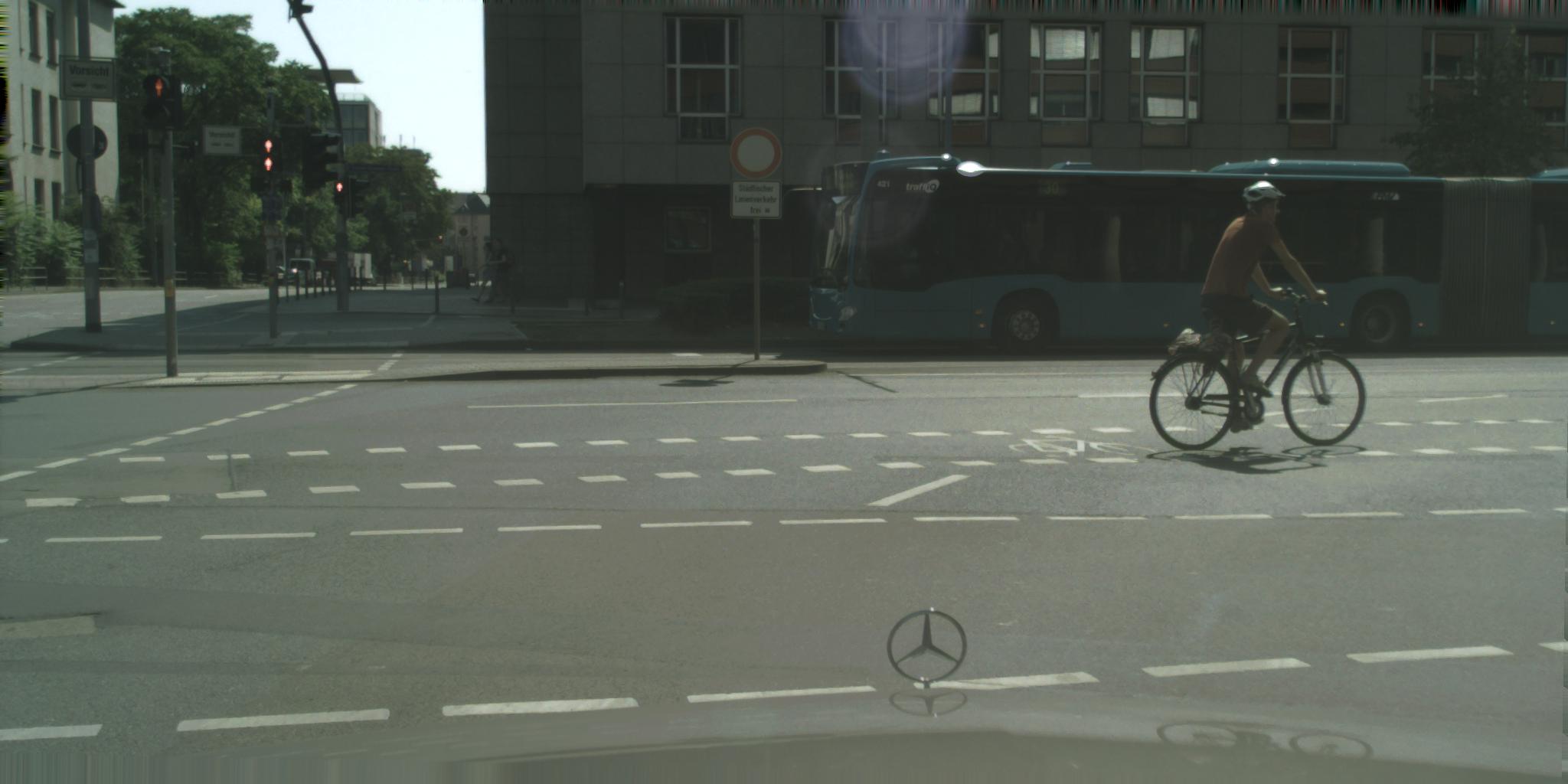} \hfill 
  \AddImg{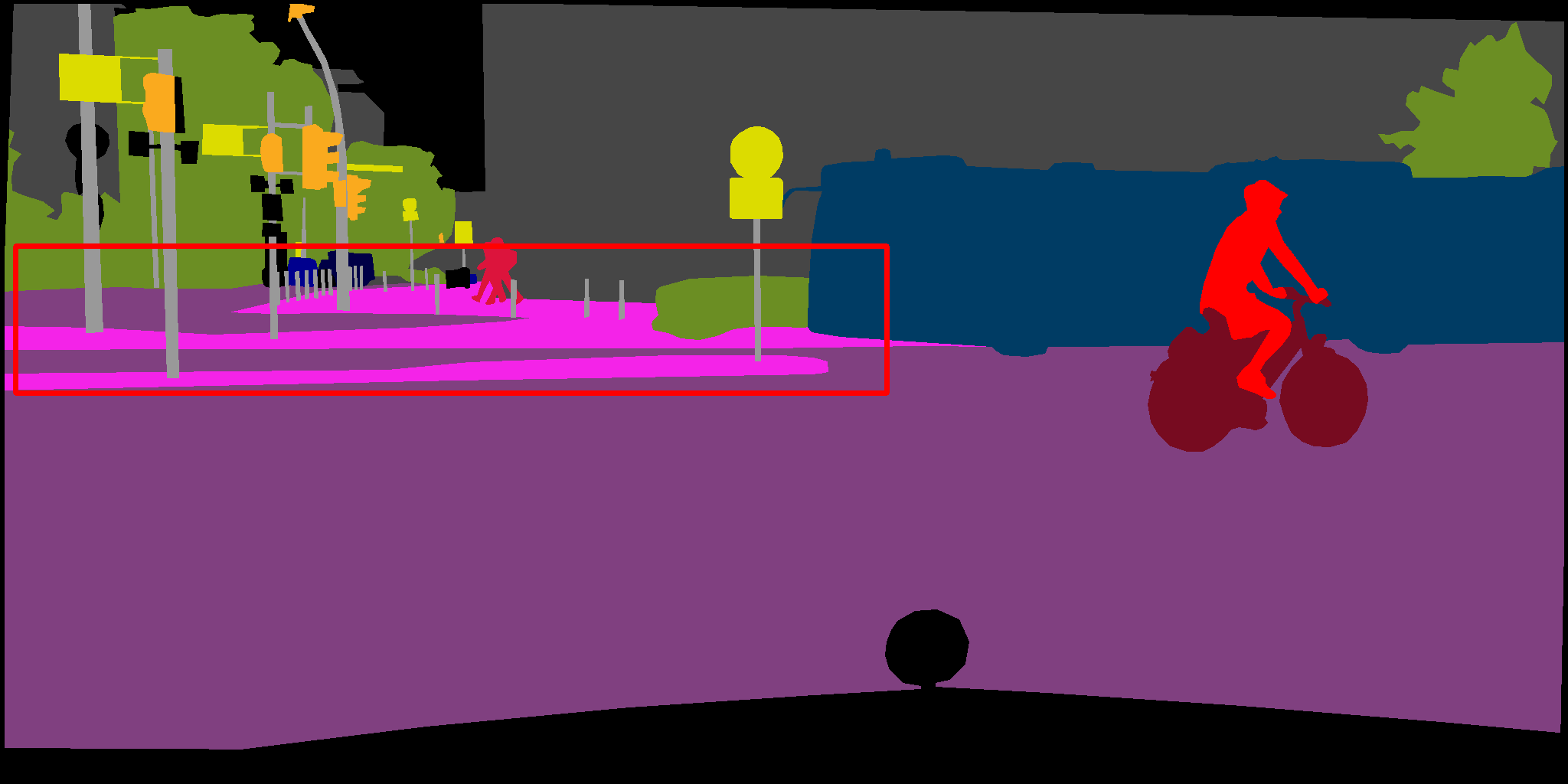} \hfill
  \AddImg{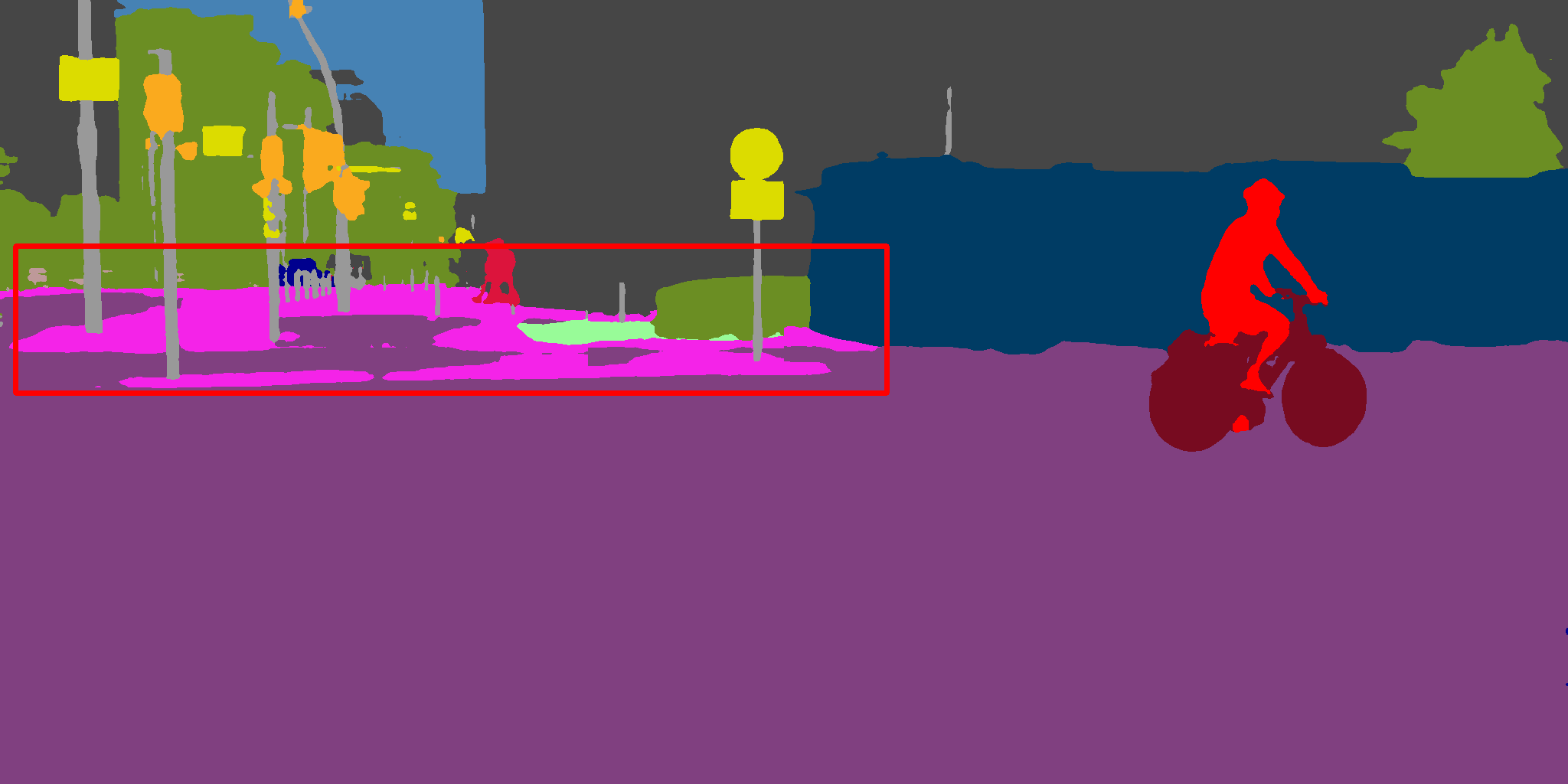} \hfill 
  \AddImg{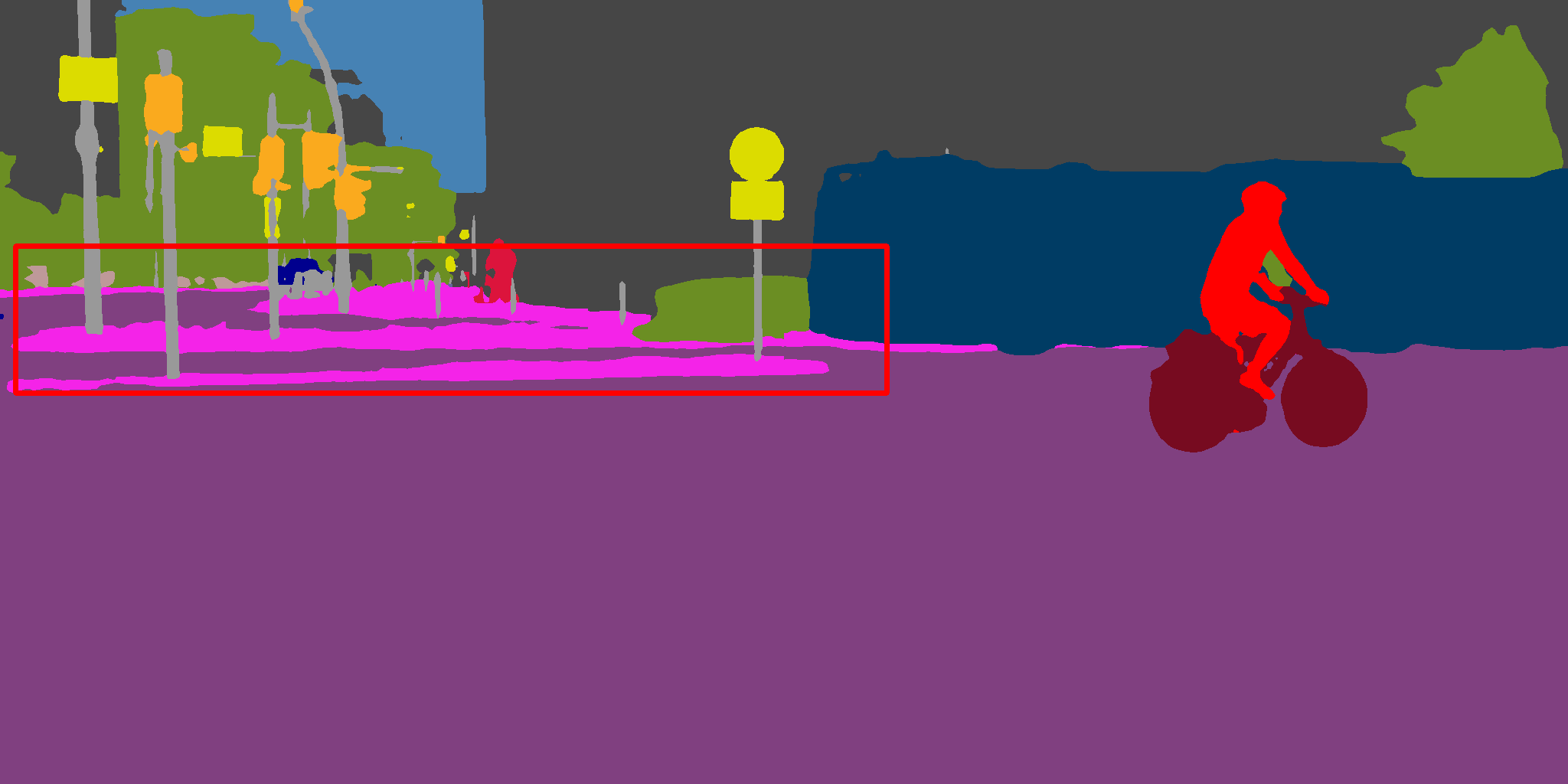}
  \\ \vspace{0.04in}
  \AddImg{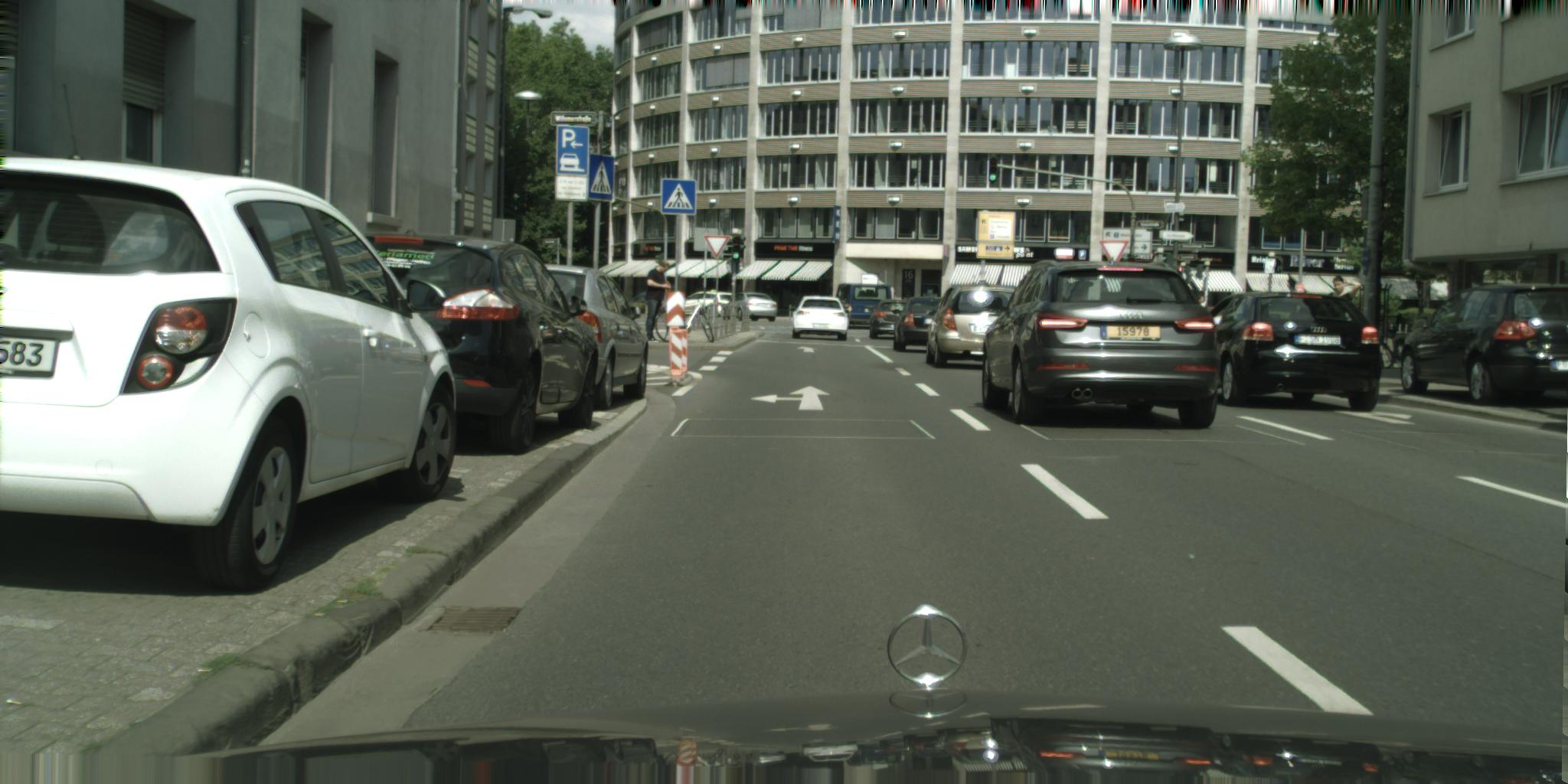} \hfill 
  \AddImg{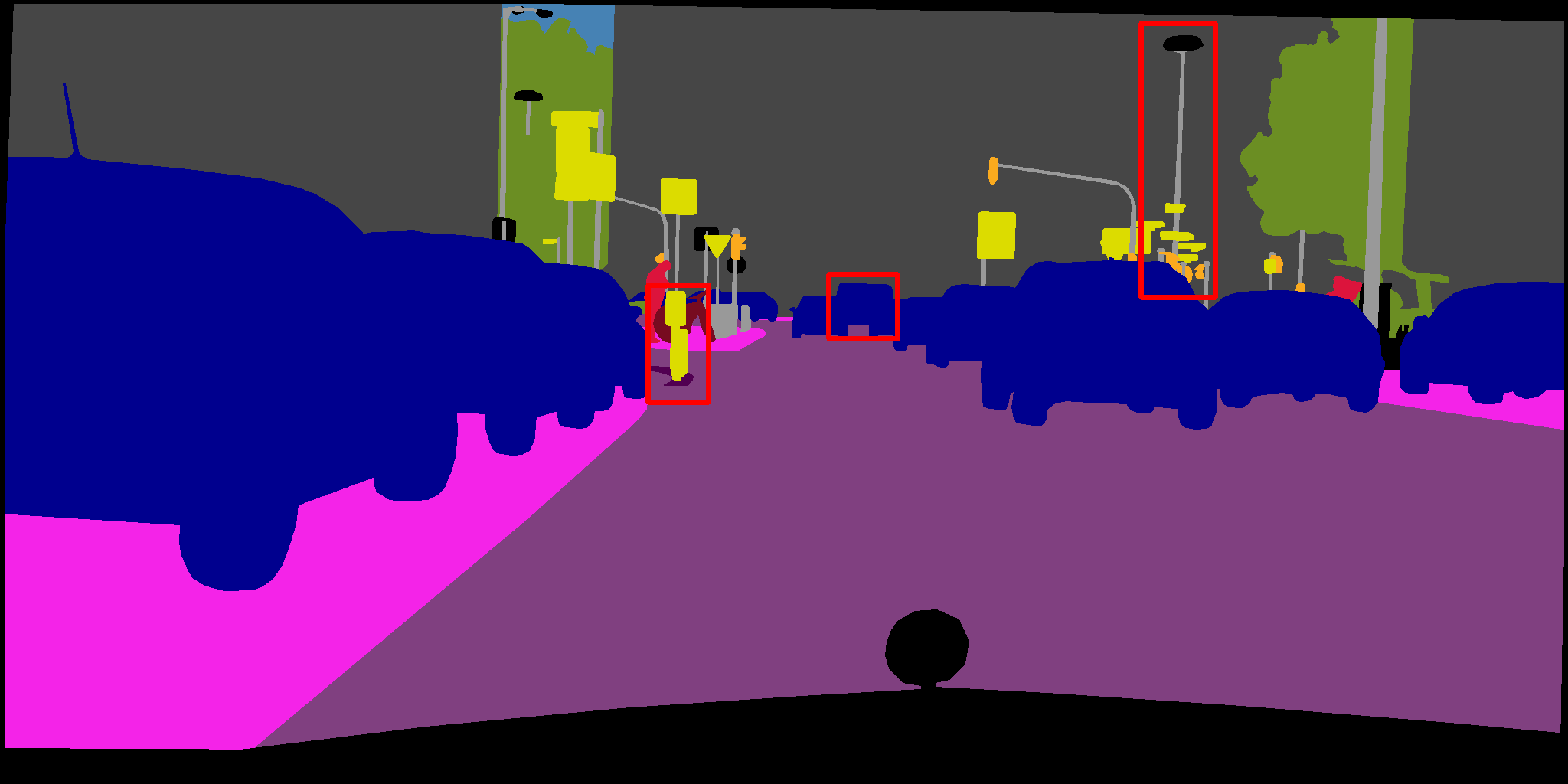} \hfill
  \AddImg{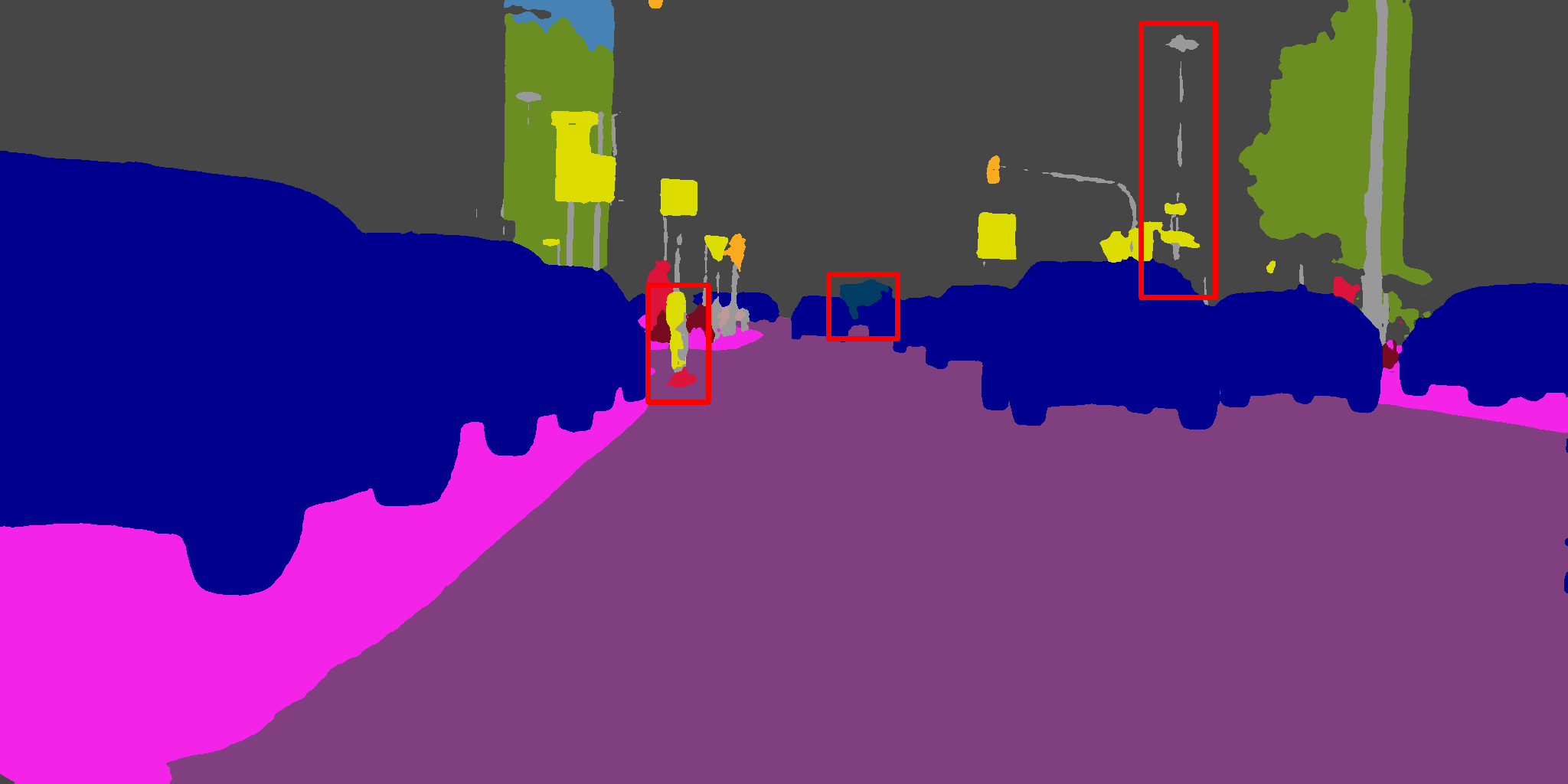} \hfill 
  \AddImg{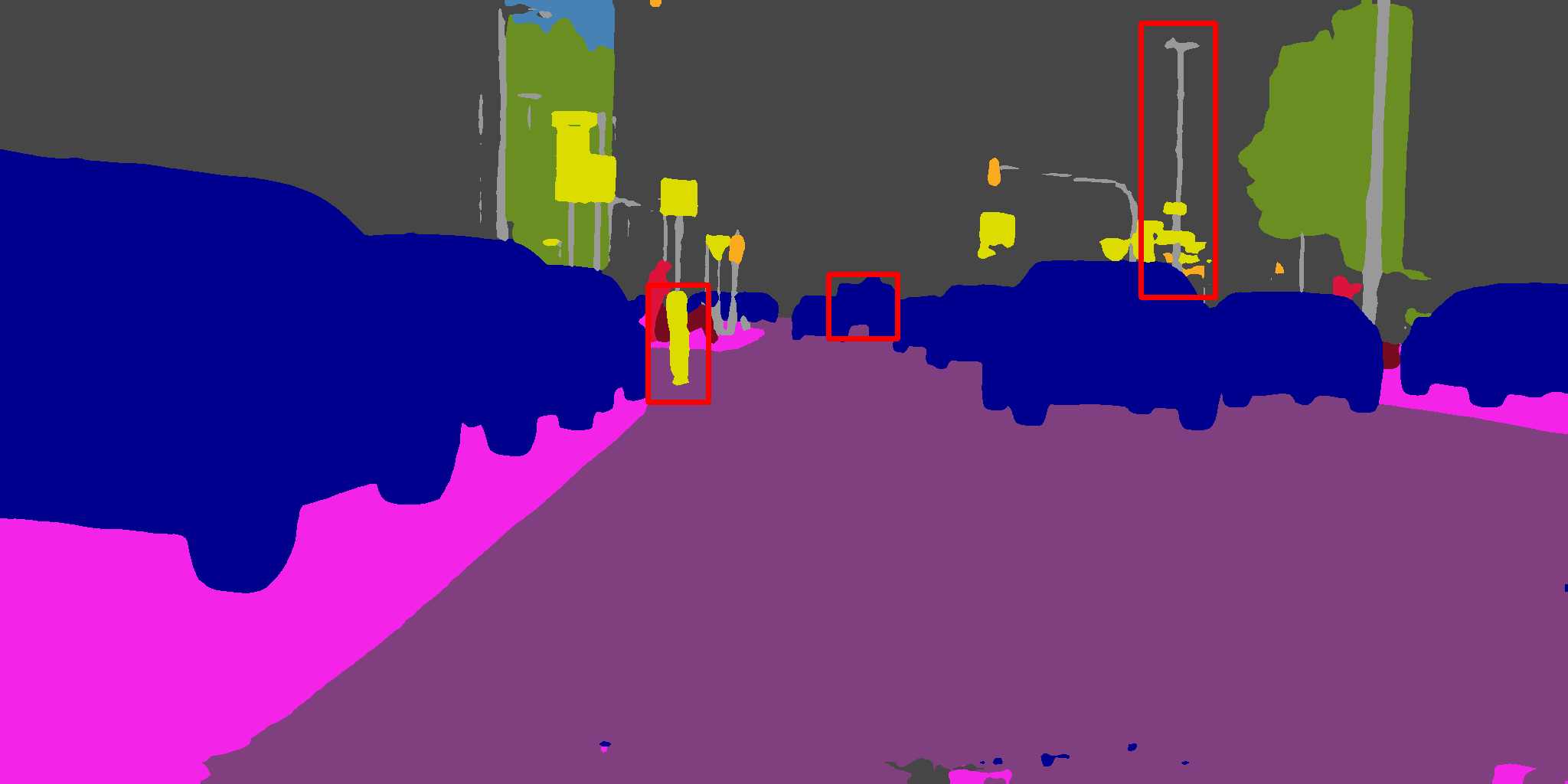}
  \caption{\textbf{Qualitative Visualization on Cityscapes val set.} The figures from left to right are input images, ground truth, segmentation maps of SegFormer\cite{xie2021segformer}, segmentation maps of our LRFormer. The significant improvements are indicated by red boxes on segmentation maps.}
  \label{fig:vis_cityscapes}
\end{figure*}

\subsection{Ablation Study}

In the following part, we conduct several ablation studies to analyze  our LRFormer.
Except for specifically mentioning, we use the following settings.
LRFormer-S is set as the baseline and trained using 8 GPUs for both classification and semantic segmentation.
For classification, our network is trained for 100 epochs in the ImageNet-1K \cite{russakovsky2015imagenet} dataset.
For semantic segmentation, our network is trained for 80K iterations in the ADE20K \cite{zhou2017scene} dataset.
Other settings are kept same as the setup in \secref{sec:setup}.

\para{Fixed pooled size.}
We reported the results in \tabref{tab:pool_size} for ADE20K semantic segmentation.
For each basic block, the pooling operation will be omitted if the feature map size is smaller than the desired pooled size.
Default fixed pooled size $m$ is $16^2$ for semantic segmentation.
Results show larger pooled size ($m \ge 16^2$) achieves saturated performance.
The default setting only introduces 5\% training memory overhead and FLOPs compared with the pooled size of $8^2$ for semantic segmentation.
Further decreasing the pooled size to $4^2$ will not introduce significant gain on improving efficiency.
When increasing the pooled size to $32^2, 48^2, 64^2$, we obtain a minor improvement or even decreased performance on ADE20K semantic segmentation.
We also observe that the FLOPs and training memory overhead are much more significant (26\% $\sim$ 170\%) when the pooled size is larger than $16^2$.
We then conduct similar experiments on Cityscapes dataset, which has a much larger input size (1024$\times$1024). The mIoU results of LRFormer-L for pooled size of $16^2$ and $32^2$ are both 83.2\%, showing that the default pooled size can also work well on larger input size. 
While adjusting pooled size based on input resolution could preserve more detail for small objects, our fixed-size design works well across all object scales. 
The high-resolution DWConv branch and multi-level feature aggregation effectively maintain small object information despite the spatial reduction.
Furthermore, ideal backbone design paradigms \cite{he2016deep,liu2022convnet,liu2021swin,wang2021pyramid} suggest basic blocks maintaining consistent settings (same pooled size in our network) across different stages. This architectural simplicity enhances implementation efficiency and reduces the effort of finding optimal parameters for each stage individually.
Considering the performance, FLOPs, training memory, and ideal backbone design paradigm,  we use fixed low-resolution pooled size as the default setting.

\para{Locality capturing.}
Our LRSA only computes the attention in low-resolution space.
Introducing spatial locality, $3\times 3$ depth-wise convolution, to our network is beneficial for getting fine-grained semantic maps. 
In \tabref{tab:exp_locality}, we analyzed the effect of the two depth-wise convolution before LRSA and in FFN.
The number of GFLOPs for these three configurations is very similar so it is not reported.
We can observe that the ADE20K performance of our \ourmethod~is improved by 0.5\% and 1.4\% and  when adding the depth-wise convolution before LRSA in FFN, with 5\% and 24\% training memory overhead. 
If both depth-wise convolutions are removed, the mIoU performance will further drop 2.4\% with 0.7GB less training memory usage, compared with the results only without the depth-wise convolution in FFN. 
This indicates that locality capturing plays a significant role in LRFormer.
Therefore, we add both of them in our LRFormer.

\begin{table}[!t]
  \centering
  \caption{
  Performance comparison between the 16×16 pooled size (original) and smaller pooled size (4$\times$4).
  }
  \label{tab:small_objects}
  \begin{tabular}{l|c|c|c|c}
  \Xhline{1pt}
  \textbf{Category} & \textbf{Metric} & \textbf{Default} & \textbf{Smaller} & \textbf{Relative Change} \\
  \Xhline{1pt}
\multirow{2}{*}{\textbf{Small}} 
& mIoU & 36.2\% & 33.7\% & -7.1\% \\
& mAcc & 46.3\% & 42.3\% & -8.8\% \\
\hline
\multirow{2}{*}{\textbf{Medium}} 
& mIoU & 48.1\% & 45.3\% & -5.7\% \\
& mAcc & 59.7\% & 56.8\% & -4.9\% \\
\hline
\multirow{2}{*}{\textbf{Large }} 
& mIoU & 57.2\% & 53.8\% & -6.0\% \\
& mAcc & 68.3\% & 64.3\% & -5.9\% \\
  \Xhline{1pt}
  \end{tabular}
  \end{table}

\para{Performance on small objects.}
To investigate how pooled size affects different-sized objects, we categorize ADE20K semantic classes into small, medium, and large categories. Discrete objects were classified based on their typical real-world dimensions, while amorphous regions (\eg, sky) were included in the large category due to their typically extensive spatial coverage in images.
Results are shown in \tabref{tab:small_objects}.
Using a smaller pooled size (4$\times$4) instead of our default setting degrades performance across all categories, with small objects experiencing the most severe impact.
This confirms that downsampling features to a resolution that is too small (\eg, 4$\times$4) can lose important semantics particularly for small objects.

\begin{table}[!t]
  \centering
  \setlength{\tabcolsep}{1mm}
  \caption{\textbf{Ablation study on the spatial locality capturing.}
  ``Memory'' is the training memory in ImageNet pretraining.
  }
  \label{tab:exp_locality}
\resizebox{\columnwidth}{!}{%
  \begin{tabular}{l|ccccc} \Xhline{1pt}
      Method & Memory & Top-1 Acc. $\uparrow$ & mIoU $\uparrow$
      \\ \Xhline{1pt}
      LRFormer-S  & 14.5GB & 81.6\% & 48.5\%  \\
      w/o DWConv (bef. LRSA) & 13.8GB  & 81.4\% & 48.0\% \\
      w/o DWConv (FFN) & 11.7GB & 81.1\% & 47.1\% \\
      w/o DWConv (Both) & 11.0GB & 80.4\% & 44.7\% \\
  \Xhline{1pt}
  \end{tabular}}  
\end{table}

\begin{table}[!t]
  \centering
  \caption{\textbf{Comparisons of our simple decoder and other popular decoder heads.}
  }
  \label{tab:exp_decoder_cmp}
  \setlength{\tabcolsep}{1mm}
\resizebox{.8\columnwidth}{!}{%
  \begin{tabular}{l|ccc} \Xhline{1pt}
      Decoder Head 
      & FLOPs $\downarrow$ & \#Params $\downarrow$ & mIoU $\uparrow$
      \\ \Xhline{1pt}
      Ours  & 40G & 32M & 49.5\% \\
      w/ OCR \cite{yuan2020object}  & 48G & 34M & 48.0\% \\
      w/ PPM \cite{zhao2017pyramid}  & 82G & 44M & 48.4\%  \\
      w/ DA \cite{fu2019dual}  & 94G & 42M & 48.9\% \\
      w/ CC \cite{huang2023ccnet}  & 84G & 42M & 48.6\% \\
  \Xhline{1pt}
  \end{tabular}
  }
\end{table}

\para{Comparions of different decoder heads.}
Our decoder head aims to predict the semantic maps from multi-level feature maps effectively and efficiently with LRSA.
The validate the LRSA of our decoder head, we compare it with several popular decode heads.
These popular decoder heads are designed for CNNs, whose output feature maps are usually 1/8 of the original image.
However, the backbone encoder of our LRFormer can output features of the 1/32 of the original image.
To make a fair comparison, we first upsample features of the last stages and concatenate them together.
Then, we feed them to the popular decoder heads. Other processes keep unchanged in these popular decoder heads.
\tabref{tab:exp_decoder_cmp} summarized the results on ADE20K semantic segmentation. The backbone is pretrained for 300 epochs on ImageNet-1K.
Compared with PPM \cite{zhao2017pyramid}, DA \cite{fu2019dual}, and CC \cite{huang2023ccnet}, our LRFormer achieves 1.1\%, 0.6\%, and 0.9\% improvement, respectively, with only fewer than 50\% FLOPs.
Compared with OCR \cite{yuan2020object}, our LRFormer obtains 1.5\% performance gain, with 83\% FLOPs.
Therefore, our default setting is more efficient and effective than other popular decoder heads.

\para{Bilinear interpolation.}
Although the self-attention is computed in a low-resolution manner, 
a bilinear interpolation is needed to fit the size as requested by the residual connection. However, we find that using LRFormer-S with an input size of 512$^2$, the bilinear interpolation only has a latency of 0.1ms, constituting a negligible 0.8\% of the overall network's latency.

\para{Decoder head with other backbones.} 
In this part,
we conduct an experiment on SegFormer-B2 replacing with our decoder. We find that SegFormer-B2 with our decoder achieves 47.3\% mIoU, 
0.8\% better than the initial SegFormer-B2 with less 28 GFLOPs. 
Thus, replacing SegFormer-B2's decoder with ours can introduce a significant improvement in terms of the performance and efficiency.

\begin{table}[!t]
  \centering
  \setlength{\tabcolsep}{3mm}
  \caption{\textbf{Discussions on the dimensions of the decoder.
  }
  When the dimension of the decoder is larger than 384, the performance will be saturated or even decreased.
  }
  \label{tab:exp_decoder_dim}
\resizebox{.85\columnwidth}{!}{%
  \begin{tabular}{c|cccc} \Xhline{1pt}
      Dimension 
      & FLOPs $\downarrow$ 
      & \#Params $\downarrow$ & mIoU $\uparrow$
      \\ \Xhline{1pt}
      128 & 27G & 30M & 47.8\%\\
      256 & 32G & 31M & 49.2\% \\
      384 & 40G & 32M & 49.5\% \\
      512 & 50G & 37M & 49.6\% \\ 
      768 & 79G & 46M &  49.2\%\\
      1024 & 117G & 58M & 49.2\%\\
  \Xhline{1pt}
  \end{tabular}} 
\end{table}

\begin{table}[!t]
  \centering
  \setlength{\tabcolsep}{0.5mm}
  \vspace{1pt}
  \caption{
  \textbf{Analysis of the memory usage and FLOPs for different input size.}
  ``Att. FLOPs'' indicates the summation of MHSA and upsampling operations.
  "Memory" is the training memory for semantic segmentation. 
  }
  \label{tab:memory_flops}
\resizebox{\columnwidth}{!}{%
  \begin{tabular}{l|cclc} \Xhline{1pt}
    Method &  Size, Batch Size & Memory $\downarrow$ & FLOPs $\downarrow$ & Att. FLOPs $\downarrow$ \\
    \Xhline{1pt}
    \ourmethod-S &  512$\times$512, 2 & 4.2GB & 40G & 0.8G \\
    SegFormer-B2 &  512$\times$512, 2 & 7.2GB & 62G  & 3.4G  \\
    \hline
    \ourmethod-S &  1024$\times$1024, 1 & 5.7GB & 145G & 0.9G \\
    SegFormer-B2 &  1024$\times$1024, 1 & 18.8GB & 279G & 54.0G \\
    \hline
    \ourmethod-S &  1536$\times$1536, 1 & 15.3GB & 319G & 1.1G \\
    SegFormer-B2 &  1536$\times$1536, 1 & OOM & 802G  & 293.6G \\
  \Xhline{1pt}
  \end{tabular}
  }
\end{table}

\para{Dimensions of the decoder head.}
To optimize the performance and computational cost of the decoder head, we employ a 1$\times$1 convolution to reduce the dimension of the concatenated multi-level features before feeding them into the decoder.
We conducted experiments with various dimension settings and compared their results in \tabref{tab:exp_decoder_dim}.
The backbone is pretrained for 300 epochs on ImageNet-1K.
The experiments show that a dimension setting of 512 achieves the best performance. However, setting the dimension to 384 results in only a 0.1\% drop in mIoU performance, while saving 25\% FLOPs. Therefore, we set the dimension of the decoder head to 384 in our LRFormer-S, reflecting the optimal trade-off between performance and computational cost.

\para{Memory and FLOPs.}
Our LRSA has a very low computational complexity of only $O(C^2+CN)$.
We numerically analyze the efficiency of our LRFormer for different input sizes, 
as well as the comparisons with the representative method SegFormer \cite{xie2021segformer}.
The analyzed results on FLOPs, attention FLOPs, and training memory are shown in \tabref{tab:memory_flops}.
For LRFormer, we additionally include the computational cost of upsampling operations.
Our \ourmethod-S costs much less memory and FLOPs than SegFormer-B2. 
Given input size of $1024\times 1024$,
the number of FLOPs of MHSA operations in our LRFormer is dramatically lower than (0.9G \textit{vs.} 54G) the self-attention in SegFormer.
We can also observe that when the input size is increasingly larger, the superiority of LRFormer will be more substantial.
This is because
increasing input size will only slightly increase the FLOPs of upsampling operations in our MHSA.

\begin{table}[!t]
  \centering
  \setlength{\tabcolsep}{2mm}
  \caption{
  \textbf{Comparisons with recent query-based frameworks on the ADE20K dataset \cite{zhou2017scene}.}
  The results of our method are marked as \textbf{bold}.
  Methods ended with ``+'' are the enhanced versions upgraded with Mask2Former's decoder.
  ``$^\dagger$'' indicates the result pretrained on ImageNet-22K and with a larger image size 640$\times$640.
  }
  \label{tab:exp_ade20k_advanced}
\resizebox{\columnwidth}{!}{%
  \begin{tabular}{l|c|c|c} \Xhline{1pt}
      Method
       & FLOPs $\downarrow$ & \#Params $\downarrow$ & mIoU $\uparrow$
      \\ \Xhline{1pt}      
      Mask2Former (Swin-T \cite{liu2021swin})  & 74G & 47M  & 47.7\% \\
      Mask2Former (Swin-S \cite{liu2021swin}) & 98G & 69M & 51.3\% \\
      \hline
      P2T-T+ \cite{wu2022p2t} & 56G & 31M & 48.2\% \\
      P2T-S+ \cite{wu2022p2t} & 70G & 43M & 49.6\% \\
      P2T-B+ \cite{wu2022p2t} & 109G & 74M & 52.5\% \\
      \hline
      \ourmethod-T+ (Ours) & 53G & 31M  & \textbf{49.4\%} \\
      \ourmethod-S+ (Ours) & 70G & 48M  & \textbf{51.3\%} \\
      \ourmethod-B+ (Ours) & 94G & 80M  & \textbf{53.7\%} \\
      \hline
      MaskFormer (Swin-B \cite{liu2021swin})$^\dagger$  & 195G & 102M & 53.1\% \\
      Mask2Former (Swin-B \cite{liu2021swin})$^\dagger$ & 223G & 107M & 53.9\% \\
      Mask DINO (Swin-B \cite{liu2021swin})$^\dagger$ & 265G & 110M  & 54.2\%\\
      SeMask (Swin-B \cite{liu2021swin})$^\dagger$  & 227G & 110M & 54.4\%\\
      \ourmethod-L+$^\dagger$ (Ours)  & 192G & 119M & \textbf{55.8\%} \\
      \hline
      MaskFormer (Swin-L \cite{liu2021swin})$^\dagger$  & 375G & 212M & 54.3\% \\
      Mask2Former (Swin-L \cite{liu2021swin})$^\dagger$  & 403G & 215M & 56.1\% \\
      Mask DINO (Swin-L \cite{liu2021swin})$^\dagger$  & 431G & 223M & 56.6\% \\
      SeMask (Swin-L \cite{liu2021swin})$^\dagger$  & 426G & 223M & 56.3\%\\
      \ourmethod-XL+$^\dagger$ (Ours)  & 365G & 205M & \textbf{58.1\%} \\
      \Xhline{1pt}
  \end{tabular}}
\end{table}

\subsection{Advanced LRFormer with Query-based Decoders}
\label{sec:lrformer+}

Recently, there emerged some query-based frameworks like MaskFormer series \cite{cheng2021per, cheng2022masked}. Though the decoders of them are a bit more complex than direct fusion strategy like SegFormer, they can achieve outstanding performance with transformers for semantic segmentation.
As mentioned before, LRFormer uses a direct fusion strategy following previous works, showing that a simple decoding strategy can also achieve state-of-the-art performance. 
In this part, we would like to explore the potential of LRFormer combined with query-based decoders.
We build a stronger version LRFormer+, which is the LRFormer encoder paired with the Mask2Former decoder.
We make a comparison with recent methods that applied query-based decoders, \ie, Mask2Former \cite{cheng2022masked}, Mask DINO \cite{li2023mask}, and SeMask \cite{jain2023semask}.
Since Mask DINO and SeMask only have an implementation based on larger backbone like Swin-L,
for a fair comparison, we reimplement these two methods with Swin-B backbone using their official code. 
We also build a powerful method P2T+ with the recent powerful P2T \cite{wu2022p2t} upgraded with the decoder of Mask2Former for a more comprehensive analysis.

We conduct the experiments in the ADE20K dataset, following the same experimental setup.
Results are shown in \tabref{tab:exp_ade20k_advanced}.
LRFormer+ demonstrated superior performance outshining recent query-based frameworks such as Mask2Former and Mask DINO. 
We can observe that Mask DINO$^\dagger$ is 0.2\% better than Mask2Former$^\dagger$ with additional 42G FLOPs. SeMask$^\dagger$ is a more efficient architecture, which surpasses Mask2Former$^\dagger$ by 0.5\% with 4G more FLOPs.
When comparing the upgraded LRFormer+ with a Mask2Former model utilizing the Swin-B backbone, LRFormer+ achieved 1.9\% higher mIoU than Mask2Former, despite LRFormer+ having slightly more parameters but significantly 31G lower FLOPs, indicating a more efficient architecture.  Furthermore, the implementation of Mask2Former with the innovative P2T backbone showcased enhanced capabilities, with the P2T-L variant reaching a mIoU of 52.5\%, 1.2\% better than Swin-S version of Mask2Former with similar FLOPs. 
Nonetheless, LRFormer+ still outperformed this configuration. 
For example, LRFormer-B+ is 1.2\% further better than the enhanced P2T-L+ version.

For a more intuitive analysis,
we visualized the accuracy-FLOPs comparisons of \tabref{tab:exp_ade20k} and \tabref{tab:exp_ade20k_advanced} in \figref{fig:cmp_ade20k}.
From the curve and the data points of this figure, LRFormer series achieve higher accuracy with fewer FLOPs compared to all other models, like Mask2Former \cite{cheng2022masked}, Mask DINO \cite{li2023mask}, and P2T \cite{wu2022p2t}.

\begin{table}[t]
  \centering
  \setlength{\tabcolsep}{2mm}
  \caption{Performance comparison of different backbones on the vision-language model LISA \cite{lai2024lisa} for referring segmentation.}
  \resizebox{.8\columnwidth}{!}{%
  \begin{tabular}{l|c|c}
  \Xhline{1pt}
  \textbf{Backbone} & \textbf{gIoU (\%)} & \textbf{cIoU (\%)} \\
  \Xhline{1pt}
  ViT-L \cite{dosovitskiy2021image} & 36.9 & 41.1 \\
  Swin-L \cite{liu2021swin} & 38.1 & 43.1 \\
  LRFormer-XL (Ours) & \textbf{40.9} & \textbf{45.7} \\
  \Xhline{1pt}
  \end{tabular}
  }
  \label{tab:lisa_results}
  \end{table}

\subsection{Application to Vision-Language Models}

While semantic segmentation remains fundamental in computer vision, the community has also shown growing interest in reasoning segmentation tasks.
These new tasks integrate visual perception with language understanding capabilities \cite{lai2024lisa, radford2021learning, liu2023visual, li2024clip}. These tasks, exemplified by representative works like LISA \cite{lai2024lisa}, leverage large vision-language models such as CLIP \cite{radford2021learning} and LLaVA \cite{liu2023visual} to segment objects based on textual descriptions. 
To demonstrate the versatility of our proposed LRFormer beyond semantic segmentation, we conduct experiments on referring segmentation to verify whether our backbone can enhance the performance of vision-language models.

\para{Experimental setup.}
For our evaluation, we use LISA \cite{lai2024lisa} with LLaVA-7B-v1 \cite{liu2023visual} as the baseline. 
We only replace the vision backbone of LISA \cite{lai2024lisa} with three different options: ViT-L \cite{dosovitskiy2021image}, Swin-L \cite{liu2021swin}, and our LRFormer-XL, 
while keeping all other components consistent.
Each backbone is pretrained on the COCO dataset \cite{lin2014microsoft} to ensure a fair comparison, and we maintain the same architecture for other parts of the vision branch. 
We use the official strategy \cite{lai2024lisa} to train each method and test on the ReasonSeg validation set \cite{lai2024lisa} for referring segmentation, which allows segmenting specific objects in images based on language prompts.
Following previous works \cite{kazemzadeh2014referitgame, mao2016generation, lai2024lisa}, 
we use gIoU and cIoU as evaluation metrics. More details of these metrics can refer to \cite{lai2024lisa}.

\para{Results.}
\tabref{tab:lisa_results} shows the performance comparison of different backbones. 
Our LRFormer-XL backbone significantly outperforms both ViT-L \cite{dosovitskiy2021image} and Swin-L \cite{liu2021swin} across both metrics. 
Specifically, LRFormer-XL achieves 40.9\% gIoU and 45.7\% cIoU, surpassing the ViT-L \cite{dosovitskiy2021image} backbone by 4.0\% and 4.6\% in gIoU and cIoU, and the Swin-L \cite{liu2021swin} backbone by 2.8\% and 2.6\%, respectively. 
These results validate that our LRFormer effectively captures global context while preserving fine-grained details necessary for reasoning tasks. 
The consistent performance improvements across both conventional semantic segmentation and referring segmentation highlight the versatility and potential of our architecture for various advanced vision-language applications.

\section{Conclusion}

In this paper, we presented a novel approach to semantic segmentation via introducing the low-resolution self-attention.
LRSA computes the self-attention in a fixed low-resolution space, 
regardless of the size of the input image, making the self-attention highly efficient. 
Extensive experiments (\eg, \figref{fig:cmp_ade20k}) on ADE20K \cite{zhou2017scene}, COCO-Stuff \cite{caesar2018coco} and Cityscapes \cite{cordts2016cityscapes} datasets show that LRFormer outperforms state-of-the-art models, 
suggesting that LRSA is adequate to keep global receptive field with negligible computational cost, \ie, FLOPs.
This study provides evidence for the effectiveness of LRSA and opens a new direction for future research.

\myPara{Acknowledgements.} 
This work is funded by NSFC (No. 62225604, 62176130), 
 the Fundamental Research Funds for the Central Universities 
(Nankai University, 070-63233089), and A*STAR Career Development Fund (No. C233312006),
and National Research Foundation Singapore under its AI Singapore Programme (AISG Award No: AISG2-GC-2023-007). 
Computation was partially supported by the Supercomputing Center of Nankai University. 
The computational work for this article was partially performed on resources of the National Supercomputing Centre, Singapore (https://www.nscc.sg).

\bibliographystyle{IEEEtran}
\bibliography{egbib}

\newcommand{\AddPhoto}[1]{\includegraphics[width=1in,keepaspectratio]{authors/#1}}

\begin{IEEEbiography}[\AddPhoto{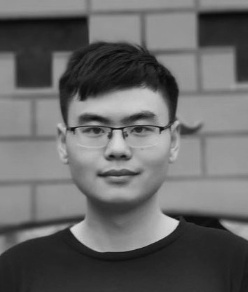}]{Yu-Huan Wu}
received his Ph.D. degree from Nankai University in 2022, advised by Prof. Ming-Ming Cheng. 
Currently, he is a research scientist at the Institute of High Performance Computing (IHPC), Agency for Science, Technology and Research (A*STAR), Singapore.
He has published 10+ papers on top-tier conferences and journals such as IEEE TPAMI/TIP/CVPR/ICCV.
His research interests include computer vision, medical imaging and autonomous driving.
\end{IEEEbiography}

\vspace{-.05in}

\begin{IEEEbiography}[\AddPhoto{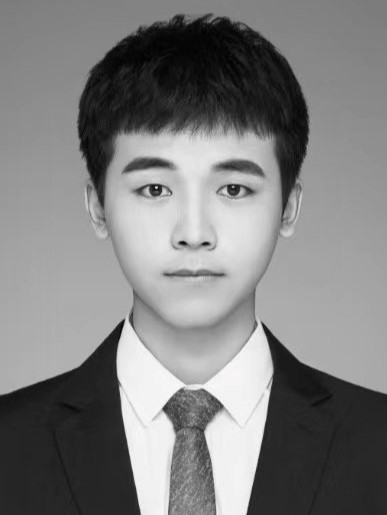}]{Shi-Chen Zhang}
received his B.E. degree in computer science
from Nankai University in 2023.
Currently, he is a Ph.D. student in Media Computing Lab, Nankai University, supervised by Prof. Ming-Ming Cheng.
His research interests include object detection and semantic segmentation.
\end{IEEEbiography}

\vspace{-.05in}

\begin{IEEEbiography}[\AddPhoto{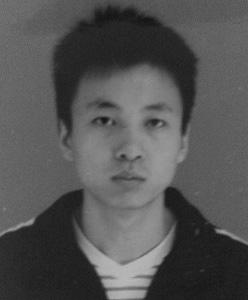}]{Yun Liu}
received his B.E. and Ph.D. degrees from Nankai University in 2016 and 2020, respectively.
Then, he worked with Prof. Luc Van Gool for one and a half years as a postdoctoral scholar at Computer Vision Lab, ETH Zurich.
Currently, he is a professor at Nankai University.
His research interests include computer vision and machine learning.
\end{IEEEbiography}

\begin{IEEEbiography}[\AddPhoto{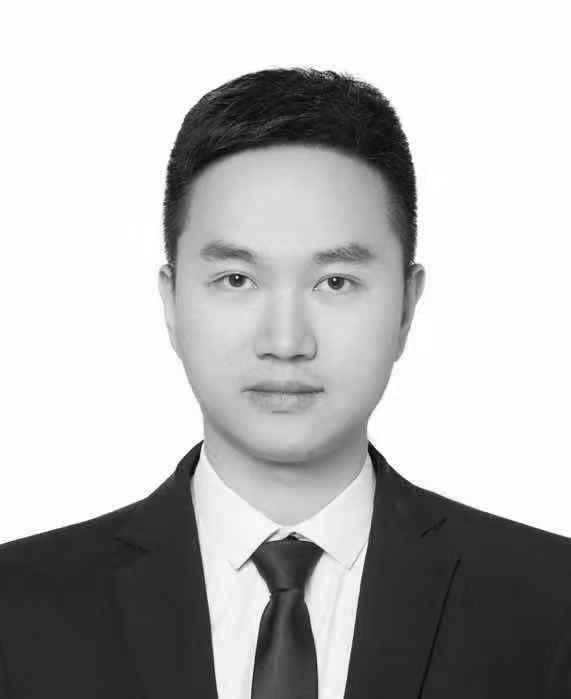}]{Le Zhang}
received his M.Sc and Ph.D.degree form Nanyang Technological University
(NTU) in 2012 and 2016, respectively.
Currently, he is a professor at UESTC.
He served as TPC member in several conferences such as AAAI, IJCAI.
He has served as a Guest
Editor for Pattern Recognition and Neurocomputing;
His current research interests include deep learning and computer vision.
\end{IEEEbiography}

\begin{IEEEbiography}[\AddPhoto{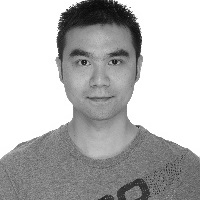}]{Xin Zhan}
received his Bachelor's (2010) and Ph.D. (2015) degrees from the University of Science and Technology of China (USTC). From 2015 to 2023, he was an Algorithm Expert at Alibaba Group, focusing on large-scale AI applications. He is currently the Algorithm Leader at Udeer AI, leading research in next-generation robotics systems. His research focuses on end-to-end autonomous driving, embodied intelligence, and multimodal vision-language-action (VLA) models.
\end{IEEEbiography}

\vspace{-.05in}

\begin{IEEEbiography}[\AddPhoto{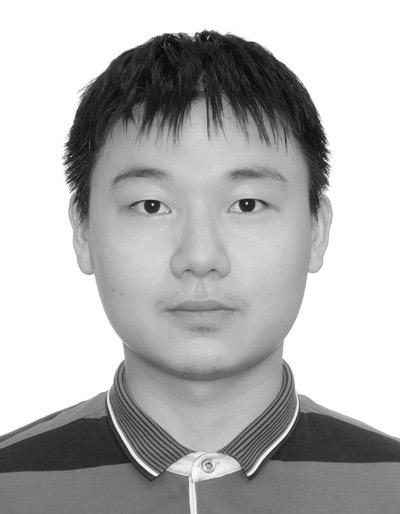}]{Daquan Zhou}
received the PhD degree from NUS, under the supervision of Prof. Jiashi Feng. 
He is currently an assistant professor of Peking University.
His research interests include deep learning, neural network compression, neural network structure design, and AutoML.
\end{IEEEbiography}

\vspace{-.05in}

\begin{IEEEbiography}[\AddPhoto{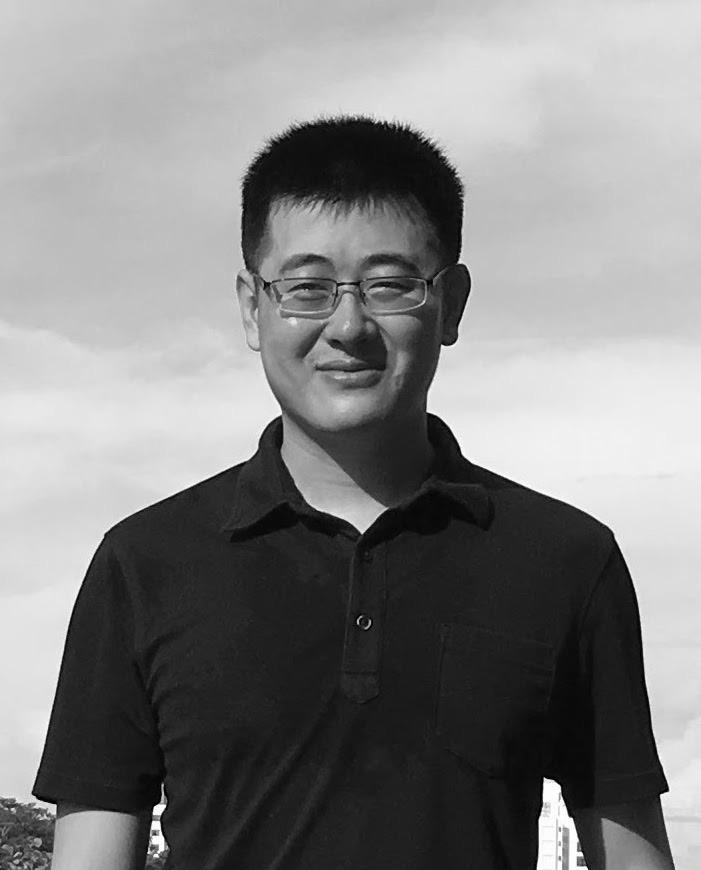}]{Jiashi Feng}
received the PhD degree from NUS, in 2014. He is currently a research lead with ByteDance. Before joining ByteDance, he was assistant professor with the Department of Electrical and Computer Engineering, National University of Singapore. His research areas include deep learning and their applications in computer vision. He received the best technical demo award from ACM MM 2012, best paper award from TASK-CV ICCV 2015, best student paper award from ACM MM 2018.
\end{IEEEbiography}

\vspace{-.05in}

\begin{IEEEbiography}[\AddPhoto{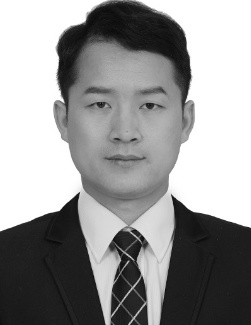}]{Ming-Ming Cheng}
received his Ph.D. degree from Tsinghua University in 2012.
Then, he did two years research fellow with Prof. Philip Torr
in Oxford.
He is now a professor at Nankai University, leading the
Media Computing Lab.
His research interests include computer graphics, computer
vision, and image processing.
He received research awards, including ACM China Rising Star Award,
IBM Global SUR Award, and CCF-Intel Young Faculty Researcher Program.
He is on the editorial boards of IEEE TPAMI/TIP.
\end{IEEEbiography}

\begin{IEEEbiography}[\AddPhoto{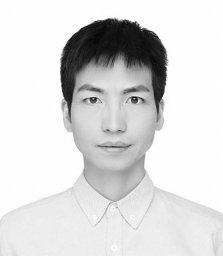}]{Liangli Zhen}
received his Ph.D. degree from Sichuan University in 2018. He is a senior scientist and group manager at the Institute of High Performance Computing (IHPC), Agency for Science, Technology and Research (A*STAR), Singapore. His research interests include machine learning and optimisation. He has led/co-led multiple projects under major national initiatives, including Singapore Aerospace Programme, AI Singapore Robust AI Grand Challenge, and AI Singapore Technology Challenge.
\end{IEEEbiography}

\end{document}